# FoMo4Wheat: Toward reliable crop vision foundation models with globally curated data

Bing Han[1, #], Chen Zhu[1, #], Dong Han[3, #], Rui Yu[1], Songliang Cao[2], Jianhui Wu[4], Scott Chapman[5], Zijian Wang[6], Bangyou Zheng[7], Wei Guo[8], Marie Weiss[9], Benoit de Solan[10], Andreas Hund[11], Lukas Roth[11], Kirchgessner Norbert[11], Andrea Visioni[12], Yufeng Ge[13], Wenjuan Li[14], Alexis Comar[15], Dong Jiang[1], Dejun Han[4], Fred Baret[1, 9], Yanfeng Ding[1, *], Hao Lu[2, *], Shouyang Liu[1, *]

# Referring to equal contributions

* Corresponding authors

Shouyang Liu: Shouyang.liu@njau.edu.cn;

Hao Lu: hlu@hust.edu.cn;

Yanfeng Ding: dingyf@njau.edu.cn

1. Engineering Research Center of Plant Phenotyping, Ministry of Education, State Key Laboratory of Crop Genetics & Germplasm Enhancement and Utilization, Jiangsu Collaborative Innovation Center for Modern Crop Production, Academy for Advanced Interdisciplinary Studies, Nanjing Agricultural University, Nanjing, China

2. National Key Laboratory of Multispectral Information Intelligent Processing Technology, School of Artificial Intelligence and Automation, Huazhong University of Science and Technology, Wuhan, China

3. Beijing University of Posts and Telecommunications, Beijing, China

4. College of Agronomy, Northwest A&F University, Yangling, China

5. School of Agriculture and Food Sustainability, The University of Queensland, Brisbane, Australia.

6. School of Electrical Engineering and Computer Science, The University of Queensland, Brisbane, Australia

7. Agriculture and Food, Commonwealth Scientific and Industrial Research Organization, Queensland Biosciences Precinct, St Lucia, Queensland, Australia

8. Graduate School of Agricultural and Life Sciences, The University of Tokyo, Tokyo, Japan

9. EMMAH UMR 1114, INRAE, Domaine Saint-Paul, Site Agroparc, Avignon, France.

10. Arvalis, LPA CAPTE, Avignon, France

11. Institute of Agricultural Sciences, ETH Zurich, Zurich, Switzerland

12. International Center for Agricultural Research in the Dry Areas (ICARDA), Rabat, Morocco

13. Department of Biological Systems Engineering, University of Nebraska-Lincoln, Lincoln, NE, US

14. State Key Laboratory of Efficient Utilization of Arable Land in China, the Institute of Agricultural Resources and Regional Planning, Chinese Academy of Agricultural Sciences, Beijing, China

15. Hiphen SAS, 22b rue Charrue, Avignon, France

## Abstract

Vision-driven field monitoring is central to digital agriculture, yet models built on general-domain pretrained backbones often fail to generalize across tasks, owing to the interaction of fine, variable canopy structures with fluctuating field conditions. We present FoMo4Wheat, one of the first crop-domain vision foundation model pretrained with self-supervision on ImAg4Wheat, the largest and most diverse wheat image dataset to date (2.5 million high-resolution images collected over a decade at 30 global sites, spanning >2,000 genotypes and >500 environmental conditions). This wheat-specific pretraining yields representations that are robust for wheat and transferable to other crops and weeds. Across ten in-field vision tasks at canopy and organ levels, FoMo4Wheat models consistently outperform state-of-the-art models pretrained on general-domain dataset. These results demonstrate the value of crop-specific foundation models for reliable in-field perception and chart a path toward a universal crop foundation model with cross-species and cross-task capabilities. FoMo4Wheat models and the ImAg4Wheat dataset are publicly available online: https://github.com/PheniX-Lab/FoMo4Wheat and https://huggingface.co/PheniX-Lab/FoMo4Wheat. The demonstration website is: https://fomo4wheat.phenix-lab.com/.

## Introduction

Agriculture is undergoing a digital transformation to sustainably address the rising global demand for food production[1–3]. While robotics may have a higher profile, this transformation first depends on the ability to accurately and efficiently monitor crops under real-world field conditions. Detailed monitoring enables data-driven precision agriculture management, which in turn reduces labor costs and mitigates environmental impacts[4,5]. Concurrently, high-throughput phenotyping across massive breeding populations accelerates the identification of superior cultivars, enhancing overall agricultural productivity[6,7]. To support these needs, camera systems mounted on ground-based platforms and drones[8] have become widely adopted due to their affordability and effectiveness to capture high-resolution crop images at spatial resolutions ranging from sub-millimeters to centimeters. These rich datasets hold considerable potential for extracting diverse crop traits, enabling comprehensive crop characterization. In particular, they support key tasks such as crop development staging (critical to agronomic applications), organ segmentation and counting, and the quantification of crop status with respect to nutrient availability, disease or pest pressures, and responses to weather and soil conditions throughout the growing season.

Deep learning-based vision models have emerged as the primary approach for addressing agricultural vision tasks[9]. Typically, these models include two main components: a backbone network and task-specific heads. The backbone is usually pre-trained on large-scale datasets comprising millions or billions of images, enabling it to learn general-purpose visual representations[10]. The learned features are then passed to task-specific heads, which are trained jointly with the backbone using labeled datasets for specific downstream tasks. Currently, most deep learning models for crop monitoring still rely on backbones pre-trained on general-domain datasets such as ImageNet[11], LVD-142M[12], and JFT-3B[13]. Although these datasets are of high quality and large quantity, they have rather limited representations of plants, crops, or agricultural environments. For instance, plant-related images are estimated to comprise only around 5% of the images in ImageNet[14].

Unlike general-domain images, crop-domain images typically exhibit densely packed, finely detailed structures, reflecting the morphological complexity of crops with overlapping leaves, stems, and reproductive organs[15]. Moreover, the variability in crop appearance is not captured: crops per se are dynamic systems; even within the same species, the appearance of crops can vary significantly due to genetic differences, growth stages, cultivation practices, and time-varying outdoor lighting[16,17]. These pronounced differences in visual patterns and structural complexity create a clear domain mismatch between crop and general-domain images, leading to features extracted from general-domain backbones being inherently suboptimal for agricultural tasks. This suboptimal representation constitutes a primary reason why many current models lack generality and fail to perform robustly under complex field conditions[18,19].

Domain-specific backbone networks trained on domain-relevant data have shown remarkable effectiveness across multiple fields, including medical imaging[20–22], remote sensing[23,24], and weather forecasting[25] (domain-specific foundation models listed in Supplementary Data 2). Such specialized backbones enable the extraction of highly representative features tailored to their respective domains. Consequently, models built on these backbones often exhibit enhanced generalization capabilities in downstream tasks and require substantially fewer labeled samples for adaptation to new applications[10]. However, modern backbone networks include millions to billions of parameters, necessitating extensive domain-specific datasets and substantial computational resources for training in new domains[26]. This challenge has driven a paradigm shift away from conventional full-parameter fine-tuning, where all backbone parameters are simultaneously updated alongside task-specific heads, towards parameter-efficient fine-tuning. In parameter-efficient fine-tuning, most backbone parameters remain frozen when adapting to downstream tasks, positioning the backbone as the central foundation determining the model performance. This shift has also reframed backbone networks as *foundation models*[10].

Although foundation models hold great promise for improving generalization in agricultural vision under real field conditions, the development of a dedicated vision foundation model for agriculture remains in its early stages[27–29]. One contributing factor is that high-throughput monitoring technologies have only recently been adopted in agriculture, and the open-access data ecosystem is still maturing. To date, the largest publicly available agricultural image dataset is iNatAg[30], with 4.7 million images representing 1,986 plant species. However, since it is primarily crowd-sourced from citizen scientists, iNatAg faces several limitations, including imbalanced species representation, variable image quality, incoherent time-series coverage, and the absence of realistic crop field conditions. The lack of large-scale, diverse, and domain-specific datasets remains a major barrier to developing an effective vision foundation model for agriculture[31].

This study investigates the performance gains in crop-related vision tasks by developing a crop-specific vision foundation model. We focus on wheat, the world's most widely cultivated and globally important food crop. This aligns with the core principle of foundation models: their effectiveness is often maximized when trained within a well-defined application domain. Thanks to our extensive global collaboration network and long-term effort in wheat image acquisition, we constructed ImAg4Wheat, the largest global wheat image dataset to date, consisting of 2.5 million high-resolution images. Using the ImAg4Wheat dataset, we developed FoMo4Wheat, the first vision foundation model tailored specifically to crop domains. We systematically evaluated FoMo4Wheat's performance across ten diverse downstream vision tasks, primarily within wheat crops to assess cross-task capabilities. Special attention was given to evaluating performance under data-scarce conditions and the model's ability to generalize to previously unseen datasets. To further examine its adaptability, we examined its cross-crop generalization potential by assessing its applicability to rice, morphologically similar to wheat, and other crops mixed with weeds. Our comprehensive evaluation underscores FoMo4Wheat's significant versatility and highlights its potential as a foundational model for various digital agriculture applications.

## Results

### Overview of ImAg4Wheat dataset and FoMo4Wheat model

ImAg4Wheat comprises 2.5 million images over 2,000 wheat genotypes cultivated under 500 distinct environmental conditions across 30 sites in 10 countries spanning a decade, covering the full crop growth cycle (Fig. 1a, Supplementary Table 5). All images were acquired in wheat breeding or experimental fields using ground-based phenotyping platforms, with ground sampling distance (GSD) ranging from 0.1 mm to 0.4 mm. Leveraging this extensive dataset, we developed the FoMo4Wheat model based on a standard Vision Transformer (ViT) architecture[32]. The model was initially pre-trained in a self-supervised manner using both Masked Image Modeling (MIM)[33] and contrastive learning strategies[34] (Fig. 1b). After freezing the backbone, FoMo4Wheat was fine-tuned following the parameter-efficient setting across 10 representative agricultural vision tasks (Fig. 1c). These tasks included six wheat-specific tasks (two at the canopy level and four at the organ level), along with two tasks on rice and two on other crops and weeds (Fig. 1d, Supplementary Table 7, Data 1). Our benchmarking yielded a unified family of FoMo4Wheat models built upon the same backbone. To support diverse deployment scenarios, we investigated three model variants, Giant, Large, and Base, comprising 1.1 B, 300 M, and 80 M parameters, respectively (Supplementary Note 2). Comprehensive benchmarking against state-of-the-art (SOTA) models across all 10 tasks shows that FoMo4Wheat consistently achieves superior performance, with particularly notable improvements on the challenging per-pixel segmentation tasks (Fig. 1e). A live demo of the FoMo4Wheat models is available at FoMo4Wheat.phenix-lab.com.

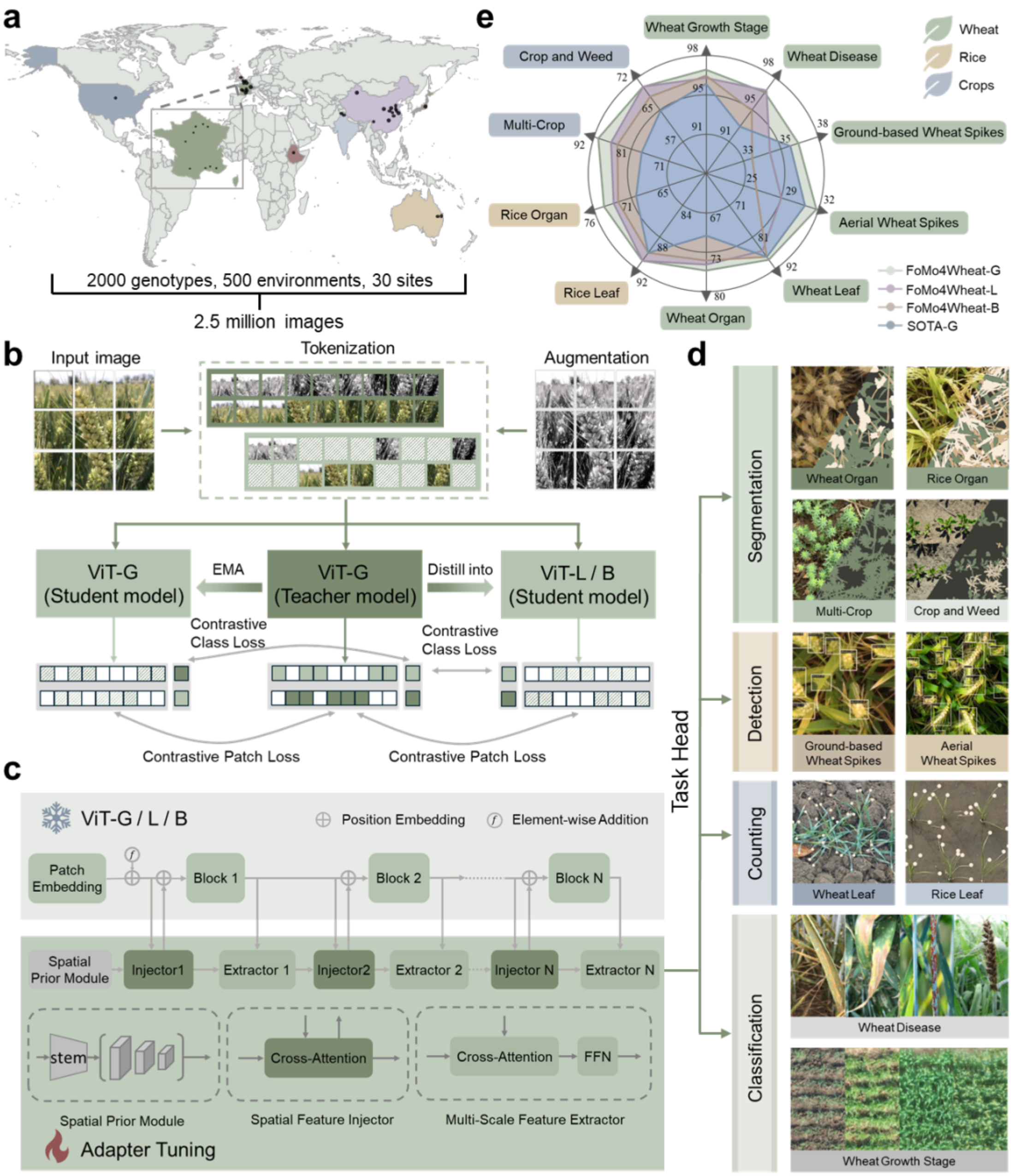

**Fig 1. Overview of ImAg4Wheat dataset and FoMo4Wheat model. a,** Global distribution of the ImAg4Wheat dataset. The dataset comprises 2.5 million high-resolution wheat images, captured across a wide range of genotype-by-environment combinations worldwide. **b,** FoMo4Wheat pre-training pipeline. The model architecture is adapted from DINOv2[12] and trained on the ImAg4Wheat dataset. Strong data augmentation is used to embed the image- and patch-level objectives into the teacher and student models. Model distillation driven by cross-entropy loss and feedback learning is further leveraged to guide the learning of the large and base models. **c,** Downstream training strategy. During adaptation to specific tasks, the pre-trained backbone is frozen, and lightweight task-specific adapters are introduced to enable efficient and flexible optimization. **d,** Overview of downstream task evaluation. FoMo4Wheat is fine-tuned with dedicated heads for multiple vision tasks, including classification (wheat growth stage and disease), detection (wheat spikes with ground-based and aerial imagery), counting (wheat and rice leaf), and segmentation (wheat and rice organ, multi-crop and crop and weed). **e,** Radar plot comparison. FoMo4Wheat consistently outperforms SOTA models across all evaluated tasks, demonstrating superior generalization and robustness across vision tasks and crop species. Each task is evaluated by its standard metric: mean average precision (mAP) for classification, average precision (AP) for detection, coefficient of determination ($R^2$, expressed as percentage) for counting, and mean Intersection over Union (mIoU) for segmentation.

## Visualization of features extracted by FoMo4Wheat model

We present and compare the visualizations of feature embeddings extracted by the FoMo4Wheat backbone and the SOTA vision foundation model DINOv2[12]. Overall, FoMo4Wheat demonstrates greater robustness and superiority to capture both high-level semantic structures and fine-grained plant features. Specifically, its embeddings form distinct clusters that clearly delineate key plant organs across various growth stages (Fig. 2a). At the development stage of 'heading' (when heads/ears/spikes have emerged from within the leaf whorl on one or more stems), FoMo4Wheat effectively distinguishes spike morphologies among diverse wheat genotypes, highlighting phenotypic variations (Fig. 2b). In disease classification, it accurately separates healthy tissues from infected regions, irrespective of symptom location (leaf, stem, or spike) and severity (Fig. 2c). When extended to crop and weed datasets, FoMo4Wheat consistently differentiates crop species from diverse weed types with high precision (Fig. 2d). In contrast, DINOv2, which presents similar emergent segmentation capabilities on general-domain images, fails to distinguish or localize meaningful crop patterns of interest.

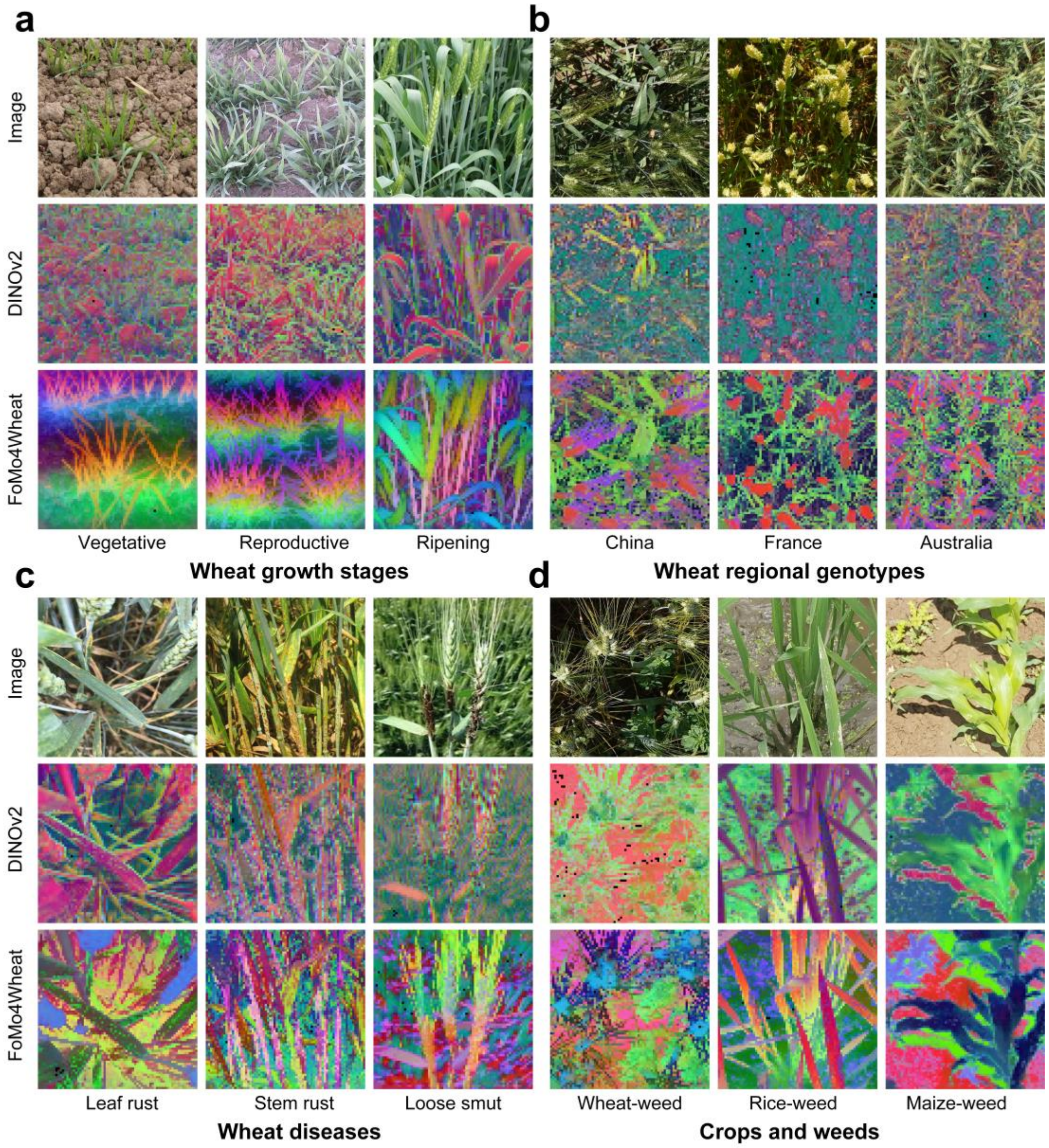

**Fig 2. Visualization of features extracted by FoMo4Wheat and DINOv2.** Feature representations extracted by FoMo4Wheat and DINOv2 are visualized using the top three principal components from Principal Component Analysis (PCA). Each subpanel displays the original image alongside the corresponding FoMo4Wheat and DINOv2 embeddings. **a**, Wheat images across growth stages: vegetative, reproductive, and ripening. **b**, Wheat images of distinct regional genotypes during the grain filling stage. **c**, Wheat images exhibiting disease symptoms on different organs, including leaf rust, stem rust, and loose smut. **d**, Multi-crop images with weed interference, covering wheat, rice, and maize.

## Canopy level classification of wheat crops

Accurate classification of wheat growth stages and diseases is essential for enabling timely and precise field management such as application of fertilizers and amelioration of weed, insect and disease challenges. We extracted feature representations for both tasks using the FoMo4Wheat and DINOv2 series and visualized the embeddings via t-SNE clustering. Compared to their DINOv2 counterparts, FoMo4Wheat variants produced more coherent clustering patterns with clearer inter-cluster boundaries in both tasks (Fig. 3b, 3d, Supplementary Fig 6, Fig7).

In the growth stage classification task (Supplementary Note 4.2), when trained on the full image dataset, all FoMo4Wheat models outperformed the SOTA (i.e., those using the same task-specific heads as FoMo4Wheat but with DINOv2 backbones; this applies hereafter). Notably, under reduced training data conditions (75%, 50%, and 25% of the original dataset), FoMo4Wheat consistently retained its performance advantage (Fig. 3a, Supplementary Table 8). The same protocol was applied to the disease classification task (Supplementary Note 4.3), where FoMo4Wheat again surpassed the SOTA models across all data regimes (Fig. 3c, Supplementary Table 8). These results collectively demonstrate that FoMo4Wheat achieves SOTA-level performance with significantly less annotated data, highlighting its superior robustness under data-scarce scenarios.

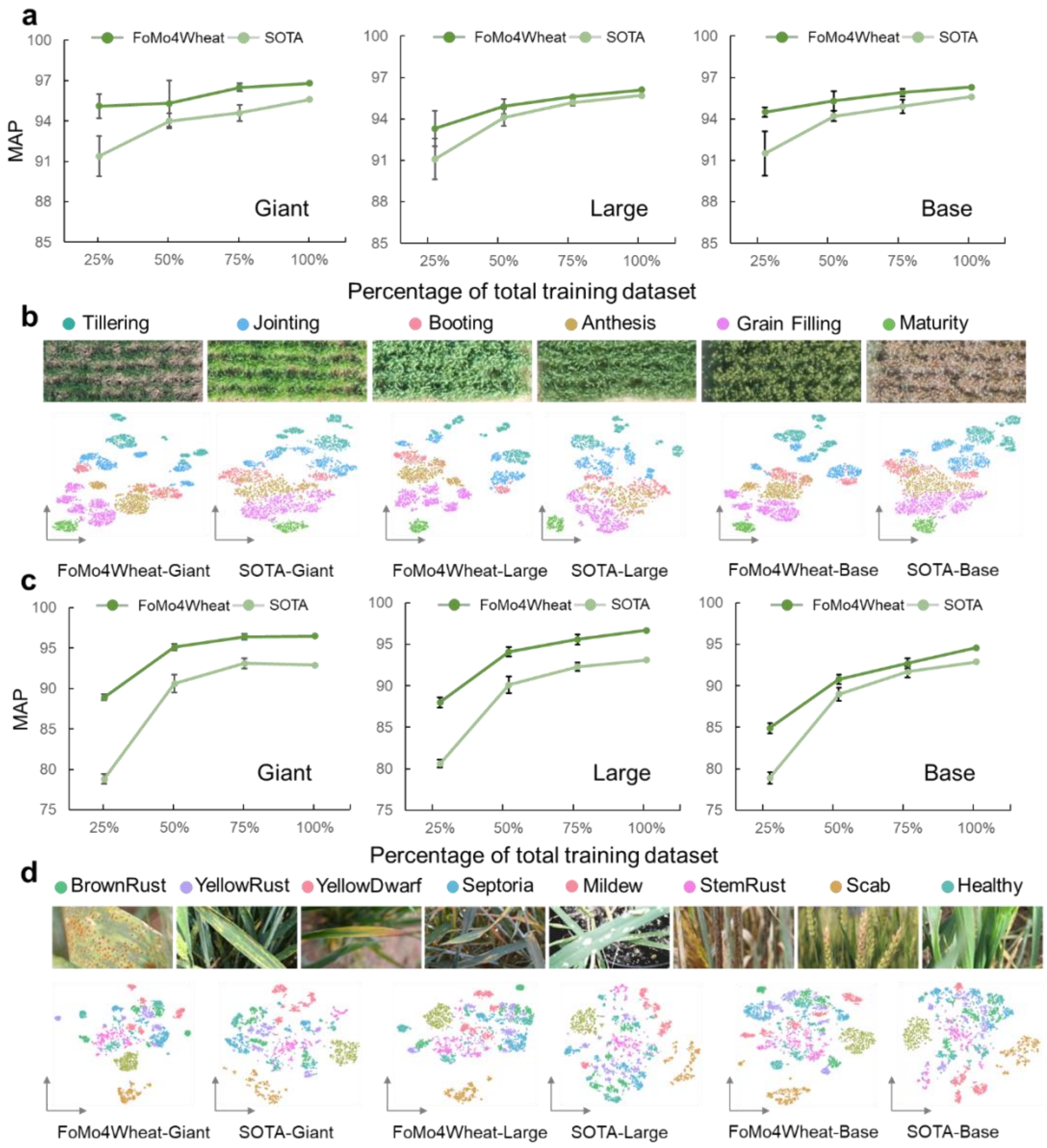

**Fig 3. Performance comparison of FoMo4Wheat and SOTA models on wheat canopy level classification tasks.** Model performance for classification is evaluated using mean average precision (mAP). **a**, Performance across varying training sample sizes for the growth stage classification task, with error bars showing variability across five random samplings for the training. **b**, 2D t-SNE visualizations of feature embeddings extracted by FoMo4Wheat and SOTA models on the growth stage dataset; colors

indicate different growth stages. **c**, Performance across varying training sample sizes for the disease classification task, with error bars representing variability across five random samplings. **d**, 2D t-SNE visualizations of feature embeddings extracted by FoMo4Wheat and SOTA models on the disease dataset; colors indicate different disease classes.

## Organ level tasks of wheat crops

Detecting wheat spikes in imagery is a critical step for quantifying spike density and estimating final yield. Both the FoMo4Wheat models and SOTA baselines were trained using ground-based images with GSD 0.4 mm (Supplementary Note 5.2). On the ground-based test set, FoMo4Wheat models slightly outperformed their SOTA counterparts (Fig. 4a, Supplementary Table 9). To evaluate robustness across acquisition platforms, we further tested the models on UAV-acquired images captured from 10 m and 12 m altitudes, corresponding to GSDs of 0.6 mm and 1.2 mm (Supplementary Note 5.3), respectively. High-GSD imagery often lacks the resolution to capture the fine structure of wheat spikes, posing substantial detection challenges. Remarkably, FoMo4Wheat models demonstrated superior performance in detecting spikes at these previously unseen GSDs, achieving more accurate localization than SOTA models (Fig. 4b, 4c, Supplementary Table 9). These results underscore the generalization capability of the FoMo4Wheat models across varying acquisition platforms and configurations, supporting scalable spike detection from ground-based systems to UAV-based monitoring.

Leaf number quantification at the emergence stage is a critical indicator of seedling vigor in cereal crops, as it directly influences tillering potential and population architecture. In the wheat leaf counting task (Supplementary Note 6.2), the FoMo4Wheat Base model outperformed the SOTA benchmark, while the Large and Giant variants yielded only marginal improvements in mean squared error (MSE) (Fig. 4d, Supplementary Table 10). Segmenting crops at the organ level is a critical prerequisite for characterizing canopy structure and further quantifying light interception, and also for organ-specific quantification of nutrient or disease stress symptoms. For organ-level segmentation of wheat images (Supplementary Note 7.2), the FoMo4Wheat models demonstrate substantial improvements over SOTA methods, with the Giant variant achieving a mean accuracy (mAcc) nearly 6 points higher, particularly excelling at delineating leaf boundaries (Fig. 4e, Supplementary Table 11).

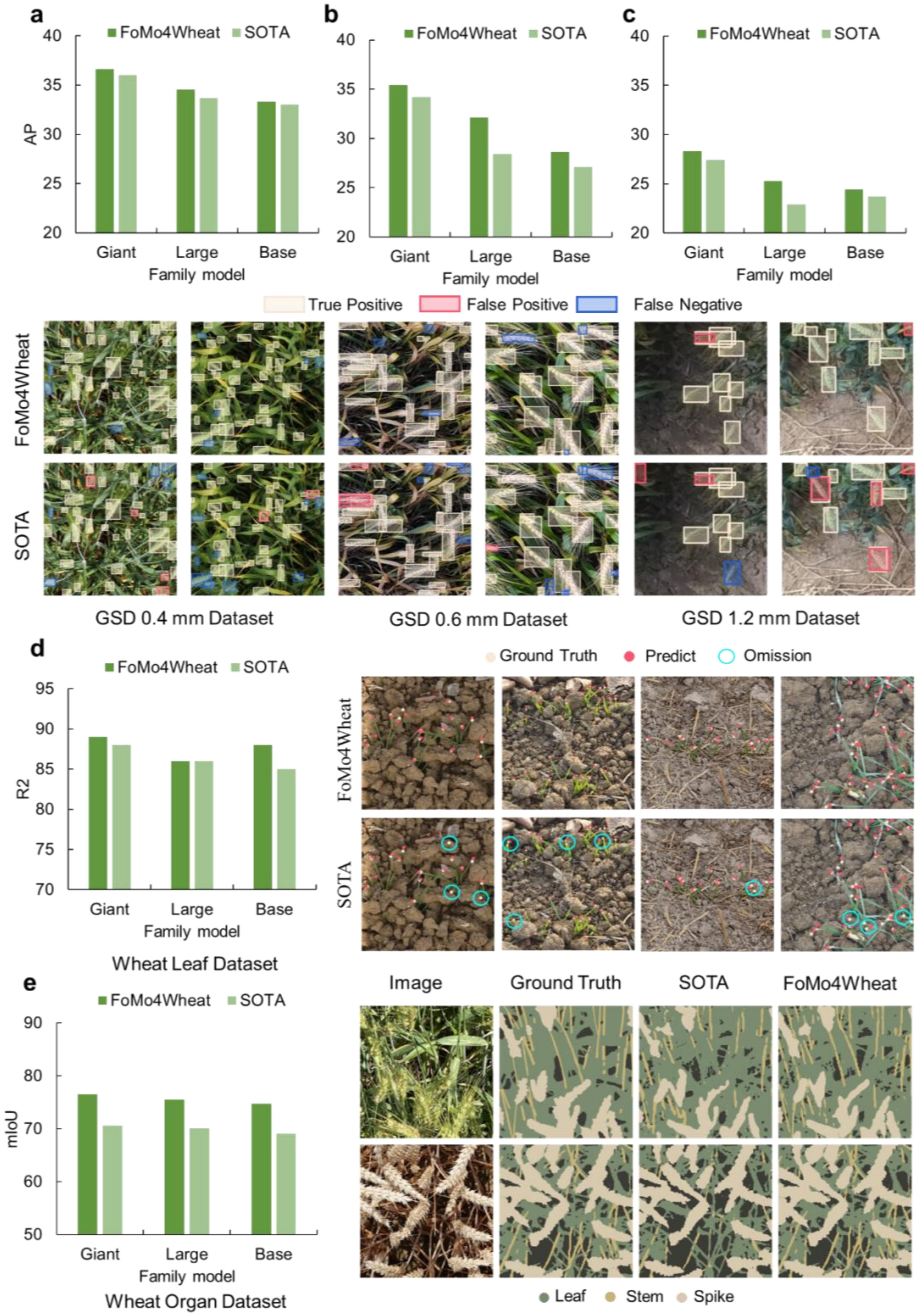

**Fig 4. Performance comparison of FoMo4Wheat and SOTA models for wheat organ level tasks.** Model performance on wheat spike detection is evaluated using Average Precision (AP) metrics. **a**, Performance on the ground-level dataset (GSD 0.4 mm). **b**, Performance on the 6 m-altitude UAV data (GSD 0.6 mm). **c**, Performance on the 10 m-altitude UAV data (GSD 1.2 mm). Detection results are compared across two examples per resolution, using ground truth, false positives, and false negatives.

**d,** Model performance on wheat leaf counting is evaluated using the coefficient of determination ($R^2$, expressed as percentage) metrics. Four example cases are used to compare the counting results, with ground truth, predict and omission. **e,** Model performance on wheat organ segmentation is evaluated using Mean Intersection over Union (mIoU) metrics. Two examples are used to highlight the segmentation of wheat spikes, stems, and leaf for FoMo4Wheat and SOTA models.

## Multi-tasks for other crops with weeds

To evaluate cross-crop generalization, we selected rice as a representative cereal crop from the Poaceae family, which shares key morphological traits with wheat[35]. For the leaf counting task (Supplementary Note 6.3), FoMo4Wheat Large and Giant only slightly outperformed SOTA methods for both wheat and rice crops (Fig. 4d; Fig. 5a; Supplementary Table 10). In contrast, in the organ segmentation task (Supplementary Note 7.3), FoMo4Wheat models achieved consistently substantial improvements over SOTA methods across both crops (Fig. 4e; Supplementary Table 11; Fig. 5b; Supplementary Table 12). These results emphasize the domain similarity between wheat and rice, despite clear visual distinctions such as the presence of water-filled backgrounds in rice paddies compared to soil-based backgrounds in wheat fields.

We further assessed model performance using a segmentation task involving a dataset comprising 9 diverse crops, including sunflower, rapeseed etc., which exhibit distinctly different structures compared to wheat (Supplementary Note 7.4). This task required accurately distinguishing vegetation from the background and correctly identifying crop species. Results indicated that, except for the SOTA Base model, both FoMo4Wheat and the larger SOTA variants achieved stronger performance in this simpler vegetation-background binary segmentation task (Fig. 5c) compared to the detailed organ segmentation for wheat and rice (Fig. 5b and Fig. 4e). Importantly, the FoMo4Wheat Giant model substantially outperformed its SOTA counterpart, with improvements exceeding 10 points in both mean Intersection over Union (mIoU) and mean accuracy (mAcc) (Fig. 5c, Supplementary Table 13).

Finally, we evaluated model generalization in mixed crop and weed scenarios involving 16 weed species across 8 crop types under real field conditions (Supplementary Note 7.5). In these challenging settings, the FoMo4Wheat and SOTA models exhibited reduced performance compared to their performance on organ-level and binary segmentation tasks. Nevertheless, the FoMo4Wheat models consistently surpassed existing SOTA methods (Fig. 5d, Supplementary Table 14). Collectively, these findings underscore the robustness and adaptability of the FoMo4Wheat models, validating their strong performance and broad generalization capabilities across diverse agricultural environments and crop species.

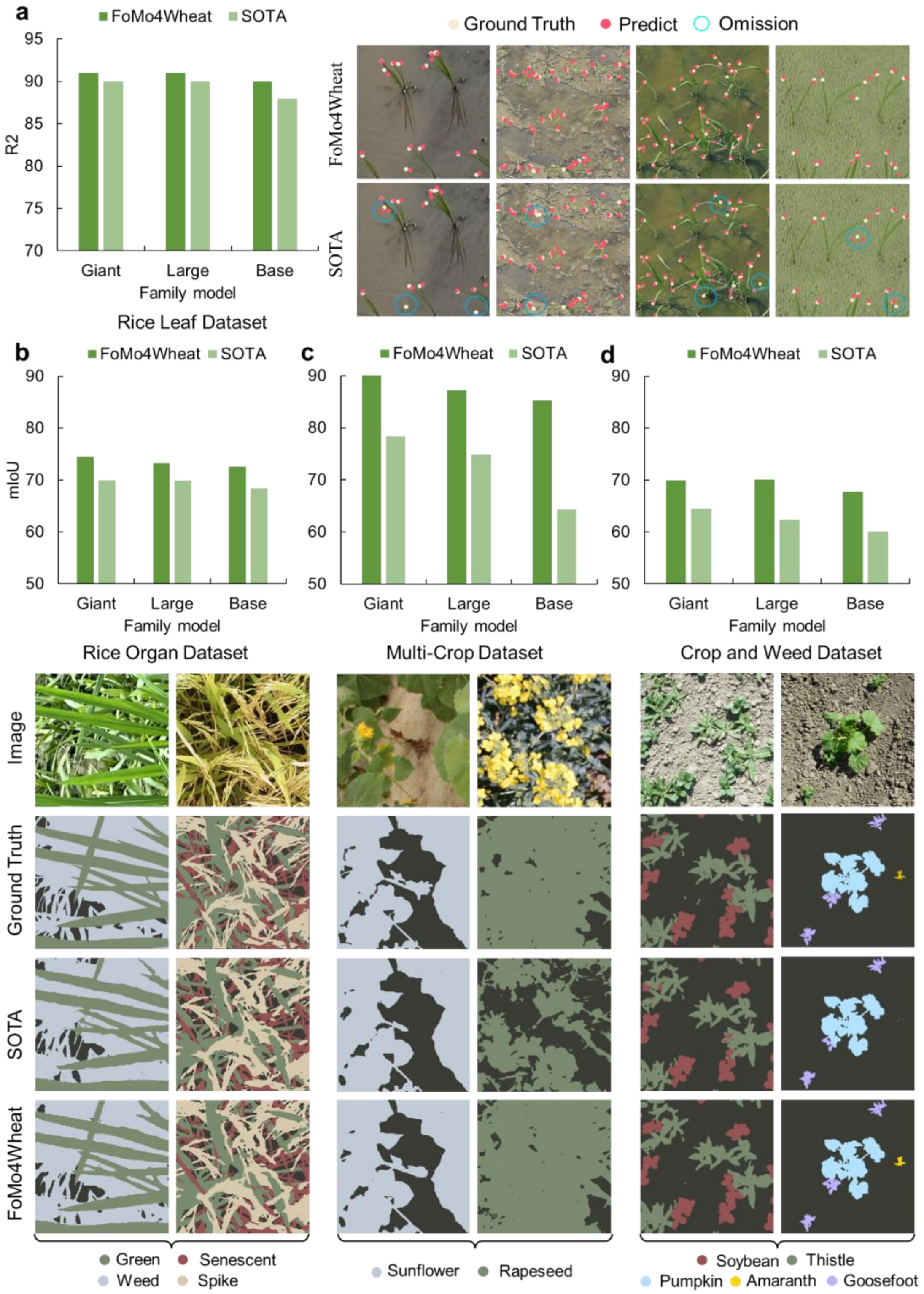

**Fig 5. Performance comparison of FoMo4Wheat and SOTA models for other crops with weeds tasks.** **a**, Rice leaf counting performance, evaluated using the coefficient of determination ($R^2$, expressed as percentage) metrics. Four example cases are used to compare the counting results, with ground truth, predict and omission. **b-d,** Model performance on segmentation tasks is evaluated using Mean Intersection over Union (mIoU) metrics. **b**, Evaluation on the Rice Organ dataset, comparing predictions using two examples, highlighting segmentation results for rice green, senescent, panicle, weed, and duckweed. **c,** Evaluation on the Multi-Crop dataset, highlighting segmentation results for rapeseed and sunflower. d, Evaluation on the Crop and Weed dataset, highlighting segmentation results for pumpkin, amaranth, goosefoot, soybean, grasses, and knotweed.

## Discussion

To the best of our knowledge, ImAg4Wheat, introduced in this study, is currently the largest and most diverse wheat image dataset, comprising 2.5 million high-resolution RGB images. Previously, the most comprehensive wheat dataset was FIP 1.0[36], which contained 153,000 curated images, organized as aligned image time series, collected over several years from a single site in Switzerland. A carefully selected subset of FIP 1.0 has been integrated into ImAg4Wheat. In contrast to large-scale online or crowd-sourced agricultural datasets such as iNatAg, which contains millions of images spanning thousands of species[30], ImAg4Wheat provides a substantially greater number of images dedicated to wheat, along with significantly higher representativeness and quality under real-world field conditions. This is the result of a globally coordinated effort involving the meticulous curation of images captured and annotated from breeding and experimental wheat fields across major wheat-producing regions. Consequently, the dataset captures extensive variability in genetic backgrounds, agronomic practices, soil types, and climatic conditions across the full wheat growth cycle (Fig. 1a, Supplementary Table 5). Moreover, ImAg4Wheat emphasizes canopy-scale imagery at high spatial resolution, clearly depicting individual plants, detailed plant organs, and soil surfaces. This targeted data collection and curation strategy ensures that the dataset is not only highly representative of wheat cropping systems, but also ideally suited for developing wheat-specific vision models optimized for real-world agricultural scenarios.

FoMo4Wheat, the first vision foundation model for wheat, surpasses SOTA models pretrained on general-domain datasets. This advantage stems primarily from its ability to learn high-level visual representations from wheat images in the ImAg4Wheat dataset, which comprehensively captures phenotypic diversity across genotypes, environments, and growth stages (Fig. 2). These enriched representations enable FoMo4Wheat to adapt to diverse downstream tasks with significantly less labeled data. For instance, in growth stage and disease classification, FoMo4Wheat achieves SOTA performance using only 30% of the training data (Fig. 3a, 3c). In organ segmentation, even without explicit task-specific training, the model's learned visual features accurately delineate organs such as spikes and leaves at the pixel level, demonstrating emergent capability. Following parameter-efficient finetuning, FoMo4Wheat achieves a 6% mIoU gain over SOTA semantic segmentation models (Fig. 4e). To rigorously assess the benefits of crop-specific foundation models, our benchmarking protocol primarily compares FoMo4Wheat and DINOv2, the leading general-domain vision foundation model, using identical task-specific heads. Beyond this head-to-head benchmark, FoMo4Wheat was also compared with top-performing models reported in recent literature. In wheat spike detection[37] and organ segmentation[38], it surpasses the best reported performances by 1.8 AP (Supplementary Table 9) and 2.83 mIoU (Supplementary Table 11), respectively.

In agricultural, the concept of foundation models has recently emerged, yet most existing efforts remain limited in capacity or training data[28,39], preventing the emergence of higher-order capabilities predicted by scaling laws[40]. As a result, they still rely on full-parameter fine-tuning rather than the parameter-efficient strategies that define modern foundation models. FoMo4Wheat addresses this gap as the first vision foundation model that fully embodies the paradigm in agriculture, achieving strong generalization across diverse wheat-related vision tasks. Remarkably, although trained exclusively on wheat imagery, it also demonstrates cross-crop generalization, performing effectively in tasks such as rice leaf counting and crop–weed segmentation. These findings suggest that visual representations learned from wheat are transferable to other crops, underscoring the potential of crop-domain foundation models to enable cross-species generalization. This trajectory paves the way toward FoMo4Crop, a universal foundation model with cross-species and cross-task capacities for major crops.

In addition, we found that initializing FoMo4Wheat with DINOv2 pretrained weights before training on the ImAg4Wheat dataset yielded better performance than random initialization (Supplementary Note 1, Table 1, Table 2). This suggests that, although plant-related images represent only a small fraction of the general-domain dataset used to train DINOv2, its representations can still benefit downstream learning on agricultural data. Nevertheless, DINOv2 alone remains suboptimal for wheat, as evidenced by degraded feature visualizations. This reflects an inherent trade-off—at the current scale of model capacity and training data—between preserving broad-domain generality and achieving strong domain-specific specialization. For wheat-focused applications, such specialization is advantageous, but the development of cross-crop foundation models such as FoMo4Crop will require carefully balancing generalization across species with crop-specific optimization. Achieving this balance will depend on new training strategies, optimized architectures, and, critically, access to much larger and more diverse datasets alongside greater computational resources. Encouragingly, the rapid expansion of high-throughput monitoring platforms is beginning to provide large-scale, multi-modal data collection across major crops, helping to make this vision achievable.

The enhanced cross-task generalization of FoMo4Wheat surpasses existing vision models, enabling highly accurate, scalable monitoring under complex field conditions. Its strong performance in organ-level semantic segmentation supports precise analysis of canopy structure and light interception[41], providing key traits for breeding. Likewise, its high accuracy in weed discrimination allows precise detection of weed patches, powering automated weeding and targeted herbicide spraying for site-specific management[42]. However, many applications require real-time, on-device inference. While the FoMo4Wheat Base model achieves competitive accuracy with a relatively compact architecture (86M parameters), its computational requirements remain heavy for edge devices. This underscores the need for advances in model compression, quantization, and knowledge distillation to deliver lightweight, field-ready models without compromising accuracy or generalization.

## Methods

**ImAg4Wheat dataset.** To enable the training of the wheat image foundation model, we aggregated a wheat image dataset of the largest scale and of the most diverse to date with significant international effort, involving 10 countries and 30 institutions worldwide, with images comprising over 2000 modern and contemporary wheat genotypes cultivated under more than 500 distinct environmental conditions, covering the full growth cycle from emergence to maturity, and spanning over a decade (2010-2024). This results in more than 2,500,000+ wheat images (Supplementary Fig 5).

**Data collection.** The wheat image data are mainly collected from non-public sources, with the exception of the FIP 1.0 data set that is public[36]. Through international collaborations, we received image data from universities and research institutions in Australia, China, Ethiopia, France, India, Japan, Switzerland, United Kingdom, and the United States. The image data integrates multi-device phenotypic data from 30 agroecological collaboration zones, which therefore establishes diverse Genotype × Environment interaction combinations, including soil types, extreme climatic conditions, and cultivation practices. Data acquisition followed standardized field trial protocols and advanced imaging technologies, employing high-throughput phenotyping devices for field plot-scale imaging, plant-scale monitoring of individual growth dynamics, multi-angle canopy-scale observations, and organ-scale imaging (leaves, stems, spikes) across developmental stages (Supplementary Table 5).

**Data preprocessing.** Since images are captured with different imaging devices and resolutions, we implemented a hybrid-scale random cropping strategy for images to maintain diversity and balance across global regions, years, and spatial scales. The image cropping sizes were constrained to a mixed range from 512×512 to 1024×1024 pixels. This enhances data loading efficiency during training and preserves original image information while retaining spatial resolution critical for canopy architecture and population characteristics.

**Downstream-task datasets.** In addition to the ImAg4Wheat dataset used for pretraining, we also leverage publicly available, self-collected, and internationally collaborated datasets tailored to six downstream wheat vision tasks, two rice vision tasks, and two generic crop vision tasks (Supplementary Table 7, Data 1). The rice- and crop-related tasks aim to justify whether the vision wheat foundation model can generalize to other crop species. The six wheat vision tasks include wheat growth stage classification, wheat disease classification, wheat head detection, UAV-based wheat spike detection, leaf tip counting, and wheat organ segmentation. The two rice vision tasks are comprised of rice leaf tip counting and rice organ segmentation. The two crop vision tasks are multi-crop segmentation and crop and weed segmentation.

### Model design

Our wheat vision foundation model implements a unified framework for different agricultural vision tasks. It mainly includes a backbone network pretrained on the ImAg4Wheat dataset and various task adapters and heads finetuned on data of downstream tasks.

**Backbone network.** DINOv2 is one of widely used frameworks for training vision foundation models in computer vision. We choose DINOv2[12] to build our backbone networks. It exploits the plain vision transformer (ViT) architecture[32] which enables seamless integration with different downstream vision tasks and allows potential extensibility towards multimodal modeling. Building upon DINOv2, we train the same counterparts of vision models specific to wheat images, referring to the FoMo4Wheat model family. This choice allows costless backbone update from the use of DINOv2 to FoMo4Wheat in the agricultural vision community.

Given a set of wheat images, FoMo4Wheat encodes images into visual representations. Concretely, for an image of size $H \times W \times 3$ is tokenized and linearly embedded to generate a sequence of patch embeddings, say, $X_p = [x_1, x_2, \ldots, x_N]^T \in R^{N\times d}$, where $N = (H/14) \times (W/14)$, and $d$ is the input embedding dimension. A learnable classification token $C \in R$ and a learnable regression token $R \in R^4$ are additionally attached on top of the patch embeddings to enhance image-level awareness, which amounts to $X'_p = [C, R, x_1, x_2, \ldots, x_N]^T \in R^{(N+5)\times d}$. Learnable positional encoding[43,44] is also applied and summed with $X'$ to generate the position-modulated embedding $X$ to enhance spatial awareness such that $X = X'_p + P$, where $P$ is the sinusoidal positional embedding[43]. This embedding is essential for dense prediction tasks that require accurate spatial localization such as wheat spike detection and plant organ segmentation. $X \in R^{T\times d}$ with $T = N + 5$ tokens is the input sequence representation to FoMo4Wheat.

At the core of FoMo4Wheat is the use of stacked Pre-LN Transformer blocks[45]. Each Pre-LN block consists of an attention sub-block and a feed-forward network (FFN) sub-block in sequence, with normalisation layers and residual connections for both sub-blocks, and outputs $X_{out} \in R^{T\times d}$ such that

$$X_{out} = \hat{X} + FFN\left(LayerNorm(\hat{X})\right), \text{where } \hat{X} = X + MHA\big(LayerNorm(X)\big).$$

Here, "$MHA$" stands for the multi-head attention, and "$LayerNorm$" is the layer normalization[46]. The MHA sub-block allows all pairwise tokens to share information between one another with self-attention weights. For any input sequence $X$, the self-attention mechanism outputs

$$Attn(X) = Softmax\left(\frac{XW_q W_k^T X^T}{\sqrt{d_k}}\right) XW_v,$$

where $W_q, W_k \in R^{d\times d_k}$, $W_v \in R^{d\times d_v}$ are trainable query, key and value parameters, respectively. In practice, self-attention is executed on $H$ different "heads" with $d_k = d_v = d/H$ to capture sophisticated relations across subspaces, which gives rise to the multi-head attention

$$MHA(X) = Concat\big(Attn_1(X), Attn_2(X), \ldots, Attn_H(X)\big)W_o,$$

where $W_o \in R^{d\times d}$ is a square projection matrix that combines different attention. Note that different heads have independent parameters. With MHA, the model is expected to capture relations within different plant organs or between individual plants at the population level, learning structural features of a plant or distribution patterns among plants. The FFN sub-block is a single hidden-layer multi-layer perceptron. The nonlinear function used has two different configurations: i) SwiGLU[47] when training with DINOv2-based pretrained initialization; ii) the standard ReLU when executing model distillation. Different numbers of transformer blocks, embedding dimensions, and numbers of heads constitute models of different capacities, i.e., FoMo4Wheat-G (1.1B params), FoMo4Wheat-L (300M params), and FoMo4Wheat-B (80M params).

**Backbone adapter and task head.** The plain ViT architecture of FoMo4Wheat has an inherent limitation: the features have limited spatial resolution due to the initial downsampling of patch tokens (1/14). Low-resolution features are sufficient for image-level vision tasks, but not for pixel-level ones. For some wheat vision tasks, such as identifying subtle disease symptoms or capturing leaf texture, high-resolution details are essential. Inspired by ViT-Adapter[48], we incorporate a similar lightweight adapter into each Transformer block for each task. The adapter reintroduces a feature pyramid of low-level visual features $\{F_1, F_2, F_3\}$ with resolutions of 1/8, 1/16, and 1/32 into the plain ViT and interacts with the Transformer block through cross-attention, thereby empowering the model with detail awareness. The adapter design is found to be beneficial for per-pixel labeling tasks such as wheat organ segmentation. To be specific, the adapter includes a spatial feature injector and a multi-scale feature extractor. By flattening and concatenating the feature pyramid into feature tokens $F_{sp}^1 \in R^{\left(\frac{HW}{8^2}+\frac{HW}{16^2}+\frac{HW}{32^2}\right)\times d}$, the spatial feature injector takes the form

$$\hat{X}^i = X^i + \gamma^i SparseAttention\left(LayerNorm(X^i), LayerNorm(F_{sp}^i)\right),$$

where $X^i$ is the input feature of the $i$-th transformer block, and $\gamma^i \in R^d$ is a learnable vector used to balance the output of the attention layer and the input feature $X^i$. Here, the attention layer $SparseAttention(\cdot)$ denotes sparse cross-attention[49]. Then, $\hat{X}^i$ is passed through the $i$-th transformer block to obtain $X^{i+1}$, and $F_{sp}^i$ is updated using the multi-scale feature extractor as

$$F_{sp}^{i+1} = \hat{F}_{sp}^i + FFN\left(LayerNorm(\hat{F}_{sp}^i)\right),$$

where $\hat{F}_{sp}^i = F_{sp}^i + SparseAttention\left(LayerNorm(F_{sp}^i), LayerNorm(X^{i+1})\right)$.

Finally, off-the-shelf task heads for different downstream tasks are incorporated at the end of the transformer block. Specifically, we adopt MLP[50] head for classification tasks (Supplementary Note 4.1, Fig 1), apply the Mask-RCNN[51] head for the detection tasks (Supplementary Note 5.1, Fig 2), the PET[52] head for counting tasks (Supplementary Note 6.1, Fig 3), and the Mask2Former[53] head for segmentation tasks (Supplementary Note 7.1, Fig 4).

## Model training

The training process of FoMo4Wheat has two stages. At the first stage, we pretrain the backbone of FoMo4Wheat on our unlabeled ImAg4Wheat dataset to learn generalizable and robust feature representations. At the second stage, we freeze the parameters of the backbone and jointly train the adapter and task heads on labeled data of different downstream tasks. The two-stage training strategy enables the use of the backbone network of FoMo4Wheat as a universal vision foundation model for wheat and parameter-efficient adaptation of the backbone to different tasks.

**Stage I: Model pretraining and distillation**. The pretraining stage of FoMo4Wheat generally follows the training recipes of DINOv2[12], which employs a self-supervised learning paradigm that synergistically combines contrastive and generative learning objectives. The key idea of contrastive learning involves generating multiple distinct "views" of the same input, e.g., through different data augmentations, and compelling the model to learn view-invariant representations. Specifically, by maximizing the similarity between the projected representations of different views of the same underlying image, the model is guided to capture intrinsic, high-level semantic properties, thereby enhancing its discriminative power. However, relying solely on contrastive learning may lead the model to overemphasize high-level semantic features while neglecting low-level, fine-grained details such as texture and color. To address this, another generative learning objective leveraging Masked Image Modeling (MIM)[33] is considered, which randomly masks a portion of the input image and tasks the model to reconstruct the masked regions. This objective incentivizes the model to learn localized and detailed visual features, being a supplement to contrastive learning.

The two learning paradigms are integrated within a knowledge distillation framework, where a momentum-updated teacher network $Teacher(\cdot)$ is used to guide the learning of a student network $Student(\cdot)$. Given inputs of a regular view and a masked view, denoted by $X_u$and $X_v$, respectively, the training process is driven by two loss functions. The first is the reconstruction loss, which takes the form

$$Loss_{rec} = \mathrm{CE}\big(Teacher(X_u)_{patch}, Student(X_v)_{patch}\big),$$

where $CE(\cdot)$ denotes the cross-entropy loss and $(X)_{patch} = [x_1, x_2, \dots, x_N]^T$ denotes the image patch tokens. It encourages the student patch representation to generate the corresponding teacher patch targets generated from the original, unmasked view of $X$. Another loss is the contrastive loss, defined by

$$Loss_{dis} = \mathrm{CE}(Teacher(X_u)_{cls}, Student(X_v)_{cls}),$$

where $(X)_{cls} = C$ is the class token. This encourages the alignment between the student and the teacher outputs.

Following this framework, we trained FoMo4Wheat-G with DINOv2 pretrained initialization. For compact variants, we employed knowledge distillation[54] to train FoMo4Wheat-L/B, defined by minimizing output divergence between models under identical inputs, instead of training from scratch (Supplementary Note 2, Table 3, Table 4). Since our training objective for FoMo4Wheat-G follows a teacher-to-student distillation paradigm, we employ an identical training pipeline where FoMo4Wheat-G serves as the teacher network and FoMo4Wheat-L/B as student networks. Crucially, the teacher network is frozen during this phase, while the student network parameters are retained as our final model. Once trained, these models could be used as vision foundation models for downstream wheat vision tasks, replacing generic large vision models trained on natural images (Detailed hyperparameter settings of different models can be found in Supplementary Table 6).

**Stage II: Task-specific model fine-tuning**. Once the backbone network of FoMo4Wheat is trained, one can freeze its parameters and only finetune the task-specific adapter and head for downstream tasks.

Considering the unique characteristics of the agricultural domain, we adjust the Sinkhorn-Knopp (SK) clustering loss from the original DINOv2 framework. Agricultural images often exhibit repetitive texture patterns such as crop fields and foliage. There often lacks well-defined foregrounds, resulting in lower inter-class discriminability such that clustering may not yield meaningful clusters and can even incur training fluctuations. Consequently, our solution is to reduce the weight of the SK loss, or even drop it in some experimental configurations, to better align with the data domain and improve training stability.

Following the practices of existing vision foundation models, we construct the FoMo4Wheat model family by training a suite of task-specific heads utilizing the features extracted by the foundation model. Particularly, during downstream task training, we freeze the parameters of the backbone, excluded from gradient updates and parameter optimization. Such a frozen backbone strategy is adopted based on two considerations: computational efficiency and model integrity. First, training only the task heads can significantly reduce computational overhead and training time. Second, freezing the backbone preserves the general-purpose feature representations, thereby mitigating the risk of catastrophic forgetting[55] during task-specific adaptation and safeguarding the model's generalization capabilities. In addition, one can safely repurpose the frozen backbone for other agricultural vision tasks without troublesome treatment of backbone finetuning.

**Model implementation.** Model training was conducted on a 6-node HPC cluster with 48 A100-80GB GPUs, utilizing distributed mixed-precision training and a multi-stage resolution strategy to balance computational efficiency and fine-grained visual learning (details of computation cost in Supplementary Note 3).

## Supplementary information

**Supplementary Information**

Supplementary Notes 1–7, Figs. 1–7 and Tables 1–14.

**Supplementary Data 1**

Task-specific evaluation datasets for foundation model validation

**Supplementary Data 2**

Details of domain-specific foundation models

## Acknowledgements

This work was Jointly supported by the National Key R&D Program of China (2022YFE0116200), the Young Scientists Fund of the National Natural Science Foundation of China (42201437 and 62106080), the Analytics for the Australian Grains Industry (UOQ2301-010OPX), the CSIRO and GRDC through project (CSP000179), the PHENOME project (ANR-11-INBS-0012), the Root2Res Project funded by HORIZON Europe (G.A. 101060124), the USDA-National Institute of Food and Agriculture (2020-68013-32371), and the Core Operational Grant for Basic Research of Chinese Academy of Agricultural Science (Y2025YC78). We also acknowledge the support from the High-performance Computing Public Platform and the Bioinformatics Center of Nanjing Agricultural University. In addition, we thank all researchers providing data support, as well as the co-authors who maintain the database and the researchers who develop analytical and predictive methods for the scientific community.

## Author information

### Authors and Affiliations

**Engineering Research Center of Plant Phenotyping, Ministry of Education, State Key Laboratory of Crop Genetics & Germplasm Enhancement and Utilization, Jiangsu Collaborative Innovation Center for Modern Crop Production, Academy for Advanced Interdisciplinary Studies, Nanjing Agricultural University, Nanjing, China**

Bing Han, Chen Zhu, Rui Yu, Dong Jiang, Fred Baret, Yanfeng Ding, Shouyang Liu

**National Key Laboratory of Multispectral Information Intelligent Processing Technology, School of Artificial Intelligence and Automation, Huazhong University of Science and Technology, Wuhan, China**

Songliang Cao, Hao Lu

**Beijing University of Posts and Telecommunications, Beijing, China**

Dong Han

**College of Agronomy, Northwest A&F University, Yangling, China**

Jianhui Wu, Dejun Han

**School of Agriculture and Food Sustainability, The University of Queensland, Brisbane, Australia.**

Scott Chapman

**School of Electrical Engineering and Computer Science, The University of Queensland, Brisbane, Australia**

Zijian Wang

**Agriculture and Food, Commonwealth Scientific and Industrial Research Organization, Queensland Biosciences Precinct, St Lucia, Queensland, Australia**

Bangyou Zheng

**Graduate School of Agricultural and Life Sciences, The University of Tokyo, Tokyo, Japan**

Wei Guo

**EMMAH UMR 1114, INRAE, Domaine Saint-Paul, Site Agroparc, Avignon, France.**

Marie Weiss, Fred Baret

**Arvalis, LPA CAPTE, Avignon, France**

Benoit de Solan

**Institute of Agricultural Sciences, ETH Zurich, Zurich, Switzerland**

Andreas Hund, Lukas Roth, Kirchgessner Norbert

**International Center for Agricultural Research in the Dry Areas (ICARDA), Rabat, Morocco**

Andrea Visioni

**Department of Biological Systems Engineering, University of Nebraska-Lincoln, Lincoln, NE, US**

Yufeng Ge

**State Key Laboratory of Efficient Utilization of Arable Land in China, the Institute of Agricultural Resources and Regional Planning, Chinese Academy of Agricultural Sciences, Beijing, China**

Wenjuan Li

**Hiphen SAS, 22b rue Charrue, Avignon, France**

Alexis Comar

## Contributions

B.H. and D.H. conceived the project with conceptual inputs from S.L. and H.L. S.L. and H.L. supervised all computational aspects of the study. S.L. and C.Z. established all standards and strategies for data collection. H.L., B.H. and D.H. designed all experimental analyses, which were implemented by B.H., D.H. and S.C. B.H., D.H., C.Z. and R.Y. curated the ImAg4Wheat dataset and all downstream experimental datasets. B.H., D.H. and S.C. performed the benchmarking of all methods against existing approaches. C.Z., R.Y., J.W., S.C., Z.W., B.Z., W.G., M.W., B.D.S., A.H., L.R., K.M., A.V., Y.G., W.L., A.C., D.J., D.H., F.B. and Y.D. provided the raw data for constructing the ImAg4Wheat dataset. S.L., H.L., B.H. and D.H. wrote the manuscript with input from all authors.

## Corresponding author

Correspondence to Shouyang Liu or Hao Lu or Yanfeng Ding.

## Ethics declarations

### Competing interests

The authors declare no competing interests.

## Data availability

All datasets will be available online: https://github.com/PheniX-Lab/FoMo4Wheat and https://huggingface.co/PheniX-Lab/FoMo4Wheat.

## Code availability

All models will be available online: https://github.com/PheniX-Lab/FoMo4Wheat and https://huggingface.co/PheniX-Lab/FoMo4Wheat.

The demonstration website is: https://fomo4wheat.phenix-lab.com/.

## References

1. Basso, B. & Antle, J. Digital agriculture to design sustainable agricultural systems. *Nat. Sustain.* **3**, 254–256 (2020).

2. El Jarroudi, M. *et al.* Leveraging edge artificial intelligence for sustainable agriculture. *Nat. Sustain.* **7**, 846–854 (2024).

3. Meemken, E.-M. *et al.* Digital innovations for monitoring sustainability in food systems. *Nat. Food* **5**, 656–660 (2024).

4. Bian, L. *et al.* A broadband hyperspectral image sensor with high spatio-temporal resolution. *Nature* **635**, 73–81 (2024).

5. Heuermann, M. C., Knoch, D., Junker, A. & Altmann, T. Natural plant growth and development achieved in the IPK PhenoSphere by dynamic environment simulation. *Nat. Commun.* **14**, 5783 (2023).

6. Wang, M. *et al.* Variation in TaSPL6-D confers salinity tolerance in bread wheat by activating TaHKT1;5-D while preserving yield-related traits. *Nat. Genet.* **56**, 1257–1269 (2024).

7. Welcker, C. *et al.* Physiological adaptive traits are a potential allele reservoir for maize genetic progress under challenging conditions. *Nat. Commun.* **13**, 3225 (2022).

8. Meraj, T. *et al.* Computer vision-based plants phenotyping: A comprehensive survey. *iScience* **27**, (2024).

9. Murphy, K. M., Ludwig, E., Gutierrez, J. & Gehan, M. A. Deep Learning in Image-Based Plant Phenotyping. *Annu. Rev. Plant Biol.* **75**, 771–795 (2024).

10. Bommasani, R. *et al.* On the Opportunities and Risks of Foundation Models. Preprint at https://doi.org/10.48550/arXiv.2108.07258 (2022).

11. Deng, J. *et al.* ImageNet: A large-scale hierarchical image database. in *2009 IEEE Conference on Computer Vision and Pattern Recognition* 248–255 (2009). doi:10.1109/CVPR.2009.5206848.

12. Oquab, M. *et al.* DINOv2: Learning Robust Visual Features without Supervision. Preprint at https://doi.org/10.48550/arXiv.2304.07193 (2024).

13. Dong, X. *et al.* CLIP Itself is a Strong Fine-tuner: Achieving 85.7% and 88.0% Top-1 Accuracy with ViT-B and ViT-L on ImageNet. Preprint at https://doi.org/10.48550/arXiv.2212.06138 (2022).

14. Russakovsky, O. *et al.* ImageNet Large Scale Visual Recognition Challenge. Preprint at https://doi.org/10.48550/arXiv.1409.0575 (2015).

15. Jiang, Y. & Li, C. Convolutional Neural Networks for Image-Based High-Throughput Plant Phenotyping: A Review. *Plant Phenomics* **2020**, (2020).

16. Walter, A., Liebisch, F. & Hund, A. Plant phenotyping: from bean weighing to image analysis. *Plant Methods* **11**, 14 (2015).

17. Yu, Y. *et al.* Crop/Plant Modeling Supports Plant Breeding: I. Optimization of Environmental Factors in Accelerating Crop Growth and Development for Speed Breeding. *Plant Phenomics* **5**, 0099 (2023).

18. Meraj, T. *et al.* Computer vision-based plants phenotyping: A comprehensive survey. *iScience* **27**, (2024).

19. Singh, A. *et al.* Challenges and Opportunities in Machine-Augmented Plant Stress Phenotyping. *Trends Plant Sci.* **26**, 53–69 (2021).

20. Hao, M. *et al.* Large-scale foundation model on single-cell transcriptomics. *Nat. Methods* **21**, 1481–1491 (2024).

21. Xiang, J. *et al.* A vision–language foundation model for precision oncology. *Nature* 1–10 (2025)

doi:10.1038/s41586-024-08378-w.

22. Zhao, T. *et al.* A foundation model for joint segmentation, detection and recognition of biomedical objects across nine modalities. *Nat. Methods* **22**, 166–176 (2025).

23. Hong, D. *et al.* SpectralGPT: Spectral Remote Sensing Foundation Model. *IEEE Trans. Pattern Anal. Mach. Intell.* **46**, 5227–5244 (2024).

24. Jakubik, J. *et al.* TerraMind: Large-Scale Generative Multimodality for Earth Observation. Preprint at https://doi.org/10.48550/arXiv.2504.11171 (2025).

25. Bi, K. *et al.* Accurate medium-range global weather forecasting with 3D neural networks. *Nature* **619**, 533–538 (2023).

26. Thompson, N., Greenewald, K., Lee, K. & Manso, G. F. The Computational Limits of Deep Learning. in *Ninth Computing within Limits 2023* (LIMITS, Virtual). doi:10.21428/bf6fb269.1f033948.

27. Li, J. *et al.* Foundation models in smart agriculture: Basics, opportunities, and challenges. *Comput. Electron. Agric.* **222**, 109032 (2024).

28. Shen, Y. *et al.* WeedNet: A Foundation Model-Based Global-to-Local AI Approach for Real-Time Weed Species Identification and Classification. Preprint at https://doi.org/10.48550/arXiv.2505.18930 (2025).

29. Xu, Y. *et al.* Smart breeding driven by big data, artificial intelligence, and integrated genomic-enviromic prediction. *Mol. Plant* **15**, 1664–1695 (2022).

30. Jain, N., Joshi, A. & Earles, M. iNatAg: Multi-Class Classification Models Enabled by a Large-Scale Benchmark Dataset with 4.7M Images of 2,959 Crop and Weed Species. Preprint at https://doi.org/10.48550/arXiv.2503.20068 (2025).

31. Mehrabi, Z. *et al.* The global divide in data-driven farming. *Nat. Sustain.* **4**, 154–160 (2021).

32. Dosovitskiy, A. *et al.* An Image is Worth 16x16 Words: Transformers for Image Recognition at Scale. Preprint at https://doi.org/10.48550/arXiv.2010.11929 (2021).

33. Zhou, J. *et al.* iBOT: Image BERT Pre-Training with Online Tokenizer. Preprint at https://doi.org/10.48550/arXiv.2111.07832 (2022).

34. Chen, T., Kornblith, S., Norouzi, M. & Hinton, G. A Simple Framework for Contrastive Learning of Visual Representations. Preprint at https://doi.org/10.48550/arXiv.2002.05709 (2020).

35. Gao, Y. *et al.* Bridging real and simulated data for cross-spatial- resolution vegetation segmentation with application to rice crops. *ISPRS J. Photogramm. Remote Sens.* **218**, 133–150 (2024).

36. Roth, L. *et al.* The FIP 1.0 Data Set: Highly Resolved Annotated Image Time Series of 4,000 Wheat Plots Grown in Six Years. 2024.10.04.616624 Preprint at https://doi.org/10.1101/2024.10.04.616624 (2025).

37. David, E. *et al.* Global Wheat Head Detection 2021: An Improved Dataset for Benchmarking Wheat Head Detection Methods. *Plant Phenomics* **2021**, 2021/9846158 (2021).

38. Wang, Z. *et al.* The Global Wheat Full Semantic Organ Segmentation (GWFSS) Dataset. *Plant Phenomics* 100084 (2025) doi:10.1016/j.plaphe.2025.100084.

39. Al Nahian, M. J., Ghosh, T., Sheikhi, F. & Maleki, F. Agri-FM+: A Self-Supervised Foundation Model for Agricultural Vision. in 5511–5523 (2025).

40. Wei, J. *et al.* Emergent Abilities of Large Language Models. Preprint at https://doi.org/10.48550/arXiv.2206.07682 (2022).

41. Zhu, C. *et al.* Genotype × environment × management analysis to define allometric rules between

leaves and stems in wheat. *J. Exp. Bot.* **75**, 6388–6404 (2024).

42. Upadhyay, A. *et al.* Advances in ground robotic technologies for site-specific weed management in precision agriculture: A review. *Comput. Electron. Agric.* **225**, 109363 (2024).

43. Vaswani, A. *et al.* Attention is All you Need. in *Advances in Neural Information Processing Systems* vol. 30 (Curran Associates, Inc., 2017).

44. Devlin, J., Chang, M.-W., Lee, K. & Toutanova, K. BERT: Pre-training of Deep Bidirectional Transformers for Language Understanding. in *Proceedings of the 2019 Conference of the North American Chapter of the Association for Computational Linguistics: Human Language Technologies, Volume 1 (Long and Short Papers)* (eds Burstein, J., Doran, C. & Solorio, T.) 4171–4186 (Association for Computational Linguistics, Minneapolis, Minnesota, 2019). doi:10.18653/v1/N19-1423.

45. He, B. & Hofmann, T. Simplifying Transformer Blocks. Preprint at https://doi.org/10.48550/arXiv.2311.01906 (2024).

46. Ba, J. L., Kiros, J. R. & Hinton, G. E. Layer Normalization. Preprint at https://doi.org/10.48550/arXiv.1607.06450 (2016).

47. Shazeer, N. GLU Variants Improve Transformer. Preprint at https://doi.org/10.48550/arXiv.2002.05202 (2020).

48. Chen, Z. *et al.* Vision Transformer Adapter for Dense Predictions. Preprint at https://doi.org/10.48550/arXiv.2205.08534 (2023).

49. Li, H. *et al.* CPSAA: Accelerating Sparse Attention using Crossbar-based Processing-In-Memory Architecture. Preprint at https://doi.org/10.48550/arXiv.2210.06696 (2023).

50. Rumelhart, D. E., Hinton, G. E. & Williams, R. J. Learning representations by back-propagating errors.

*Nature* **323**, 533–536 (1986).

51. He, K., Gkioxari, G., Dollár, P. & Girshick, R. Mask R-CNN. in *2017 IEEE International Conference on Computer Vision (ICCV)* 2980–2988 (2017). doi:10.1109/ICCV.2017.322.

52. Liu, C., Lu, H., Cao, Z. & Liu, T. Point-Query Quadtree for Crowd Counting, Localization, and More. in *2023 IEEE/CVF International Conference on Computer Vision (ICCV)* 1676–1685 (IEEE, Paris, France, 2023). doi:10.1109/iccv51070.2023.00161.

53. Cheng, B., Misra, I., Schwing, A. G., Kirillov, A. & Girdhar, R. Masked-attention Mask Transformer for Universal Image Segmentation. in *2022 IEEE/CVF Conference on Computer Vision and Pattern Recognition (CVPR)* 1280–1289 (IEEE, New Orleans, LA, USA, 2022). doi:10.1109/cvpr52688.2022.00135.

54. Hinton, G., Vinyals, O. & Dean, J. Distilling the Knowledge in a Neural Network. Preprint at https://doi.org/10.48550/arXiv.1503.02531 (2015).

55. Kirkpatrick, J. *et al.* Overcoming catastrophic forgetting in neural networks. *Proc. Natl. Acad. Sci.* **114**, 3521–3526 (2017).

# Supplementary Notes

## 1. Training from scratch or with DINOv2 pretrained weights for the Giant model

Weight initialization significantly affects the performance of the foundation model[1]. To train the FoMo4Wheat Giant model, we compared two weight initialization strategies: random initialization and initialization by DINOv2 pretrained weights. The DINOv2 pre-trained weights may offer a better starting point due to large-scale pretraining on the LVD-142M. Two multi-class crop segmentation tasks including wheat organ segmentation and crop multi-class segmentation were used for validation, because multi-class segmentation is sufficiently difficult to highlight the performance difference between different initialization strategies. The results reported in Supplementary Tables 1 and 2 demonstrated that FoMo4Wheat Giant model initialized with pretrained DINOv2 weights achieved superior performance than the model trained from scratch.

**Supplementary Table 1.** Performance of the FoMo4Wheat Giant model with different weight initialization strategies on wheat organ segmentation. The relative improvement is shown in red and blue.

| Weight Initialization | Class IoU | | | | mIoU |
|---|---|---|---|---|---|
| | background | head | stem | leaf | |
| Random | 85.81 | 83.56 | 46.85 | 82.30 | 74.63 |
| Pre-trained | 85.97 (+0.16) | 84.48 (+0.92) | 52.73 (+5.88) | 82.76 (+0.46) | 76.49 (+1.86) |
| **Weight Initialization** | **Class Acc** | | | | **mAcc** |
| | background | head | stem | leaf | |
| Random | 90.66 | 92.11 | 58.65 | 92.26 | 83.42 |
| Pre-trained | 91.80 (+1.14) | 92.80 (+0.69) | 67.74 (+9.09) | 91.11 (-1.15) | 85.86 (+2.44) |

**Supplementary Table 2.** Performance of the FoMo4Wheat Giant model with different weight initialization strategies on multi-class crop segmentation. Best performance is in **boldface**. The relative improvement is shown in red and blue.

| Weight Initialization | Class IoU | | | | | | | | | mIoU |
|---|---|---|---|---|---|---|---|---|---|---|
| | background | Wheat | Rapeseed | Maize | Sunflower | Sugarbeet | Mix | Rice | Potato | |
| Random | 87.84 | 86.97 | 94.68 | 90.45 | 91.36 | 86.10 | 78.30 | 91.17 | 88.79 | 88.41 |
| Pre-trained | 87.45 (-0.39) | 86.66 (-0.31) | 94.40 (-0.28) | 92.30 (+1.85) | 94.82 (+3.46) | 96.22 (+10.12) | 78.22 (-0.08) | 91.28 (+0.11) | 95.03 (+6.24) | 90.71 (+2.3) |
| Weight Initialization | Class Acc | | | | | | | | | mAcc |
| | background | Wheat | Rapeseed | Maize | Sunflower | Sugarbeet | Mix | Rice | Potato | |
| Random | 93.78 | 93.73 | 97.12 | 96.42 | 94.45 | 87.22 | 86.26 | 94.42 | 97.70 | 93.46 |
| Pre-trained | 93.30 (-0.48) | 92.98 (-0.75) | 97.26 (+0.14) | 95.93 (-0.49) | 97.20 (+2.75) | 97.63 (+10.41) | 88.12 (+1.86) | 94.56 (+0.14) | 97.54 (-0.16) | 94.95 (+1.49) |

## 2. Direct training or distillation for Large/Base variants

Once the FoMo4Wheat Giant model was trained, there are two possible strategies to train the subsequent Large and Base variants: i) direct training from scratch on the whole dataset or ii) distillation from the Giant model. Similarly, the two segmentation tasks (wheat organ segmentation and crop multi-class segmentation) were repurposed as benchmarks. For direct training, the Large and Base models were trained using the same protocol as the Giant model. For pretrained weight initialization, the Large and Base models were initialized with DINOv2 pretrained weights of the same model capacity. We follow the distillation pipeline described in the main text. Results demonstrated that the distilled Large/Base variants consistently outperformed their direct training counterparts (Supplementary Tables 3 and 4).

**Supplementary Table 3.** Performance of FoMo4Wheat Large/Base model with different methods on wheat organ segmentation. The relative improvement is shown in red and blue.

| Weight Initialization | Architecture | Class IoU | | | | mIoU |
|---|---|---|---|---|---|---|
| | | background | head | stem | leaf | |
| Scratch | Large | 85.29 | 81.68 | 43.28 | 81.21 | 72.87 |
| Distill | Large | 86.01 (+0.72) | 83.72 (+2.04) | 49.90 (+6.62) | 82.3 (+1.09) | 75.48 (+2.61) |
| Scratch | Base | 84.52 | 78.58 | 34.41 | 79.66 | 69.29 |
| Distill | Base | 85.90 (+1.38) | 83.31 (+4.73) | 47.67 (+13.26) | 82.21 (+2.55) | 74.77 (+5.48) |

| Method | Architecture | Class Acc | | | | mIoU |
|---|---|---|---|---|---|---|
| | | background | head | stem | leaf | |
| Scratch | Large | 91.06 | 91.15 | 54.55 | 91.02 | 81.94 |
| Distill | Large | 92.06 (+1.00) | 92.59 (+1.44) | 64.47 (+9.92) | 90.66 (-0.36) | 84.94 (+3.00) |
| Scratch | Base | 90.44 | 87.82 | 42.48 | 90.95 | 77.93 |
| Distill | Base | 91.15 (+0.71) | 92.16 (+4.34) | 60.14 (+17.66) | 91.68 (+0.73) | 83.78 (+5.85) |

**Supplementary Table 4.** Performance of FoMo4Wheat Large/Base model with different methods on the crop multi-class segmentation task. The relative improvement is shown in red and blue.

| Weight Initialization | Architecture | Class IOU | | | | | | | | | mIoU |
|---|---|---|---|---|---|---|---|---|---|---|---|
| | | background | Wheat | Rapeseed | Maize | Sunflower | Sugarbeet | Mix | Rice | Potato | |
| Scratch | Large | 85.61 | 78.36 | 81.93 | 76.07 | 78.59 | 75.54 | 68.17 | 86.38 | 75.96 | 78.51 |
| Distill | Large | 87.95 (+2.34) | 87.39 (+9.03) | 94.47 (+12.54) | 90.21 (+14.14) | 85.25 (+6.66) | 81.70 (+6.16) | 79.59 (+11.42) | 89.31 (+2.93) | 88.73 (+12.77) | 87.18 (+8.67) |
| Scratch | Base | 85.65 | 77.01 | 77.72 | 71.57 | 78.19 | 83.04 | 65.05 | 87.39 | 77.68 | 78.14 |
| Distill | Base | 87.83 (+2.18) | 85.27 (+8.26) | 94.45 (+16.73) | 90.18 (+18.61) | 85.07 (+6.88) | 76.87 (-6.17) | 76.07 (+11.02) | 89.05 (+1.66) | 82.75 (+5.07) | 85.28 (+7.14) |

| Weight Initialization | Architecture | Class Acc | | | | | | | | | mAcc |
|---|---|---|---|---|---|---|---|---|---|---|---|
| | | background | Wheat | Rapeseed | Maize | Sunflower | Sugarbeet | Mix | Rice | Potato | |
| Scratch | Large | 93.45 | 85.99 | 87.35 | 84.41 | 81.33 | 81.78 | 86.04 | 90.52 | 92.38 | 87.03 |
| Distill | Large | 93.54 (+0.09) | 94.40 (+8.41) | 96.72 (+9.37) | 92.93 (+8.52) | 90.94 (+9.61) | 88.61 (+6.83) | 88.55 (+2.51) | 92.42 (+1.90) | 97.96 (+5.58) | 92.90 (+5.87) |
| Scratch | Base | 92.96 | 84.14 | 84.83 | 80.6 | 82.54 | 84.28 | 88.06 | 93.13 | 85.26 | 86.20 |
| Distill | Base | 93.18 (+0.22) | 92.57 (+8.43) | 97.31 (+12.48) | 93.64 (+13.04) | 91.12 (+8.58) | 88.80 (+4.52) | 87.93 (-0.13) | 92.18 (-0.95) | 91.65 (+6.39) | 92.04 (+5.84) |

## 3. Implementation Details

Our experiments were conducted on an HPC cluster equipped with NVIDIA Ampere GPU infrastructure, where intra-server GPU communication utilizes NVLink and high-speed interconnects to establish a multi-node deep learning environment optimized for efficient training. We employ a mixed-precision training strategy to minimize time consumption under constrained computational resources, as half-precision operations execute on FP16 Tensor Cores achieve 2 times higher arithmetic throughput compared to TF32 on NVIDIA Ampere GPUs. Distributed data parallelism, a widely adopted training strategy for large-scale datasets on HPC clusters, significantly reduces model training duration through distributed data processing. The full model training is performed on a 6-node compute cluster, with each node housing 8 A100-80GB GPUs.

To prevent small objects from vanishing in pixel-level downstream tasks (e.g., segmentation or detection) under low-resolution conditions, we employed a multi-stage resolution scaling strategy where the model underwent sequential low-resolution to high-resolution training phases. The initial five-day training utilized a 224×224-pixel global crop size, effectively establishing core visual representations while leveraging computational advantages of lower resolution including accelerated iteration cycles, larger feasible batch sizes, and reduced memory footprints. Subsequently, to enhance fine-grained feature discrimination for pixel-level tasks, model weights were transferred to a high-resolution regime with a 518×518 image patches; this high-resolution learning phase required three additional days for model fine-tuning, which facilitates the awareness of richer spatial details in higher-resolution inputs. The multi-stage training approach strategically balances computational efficiency with high-fidelity representation learning, optimizing resource utilization without compromising downstream task performance.

## 4. Classification Tasks

***Classification Task Head.*** **We first evaluate on image-level classification tasks. Given the** FoMo4Wheat model **backbone, we follow the linear probing protocol where a linear classifier is used as the task head while the backbone parameters are kept frozen. We evaluate on two wheat classification tasks. To highlight the performance difference, we follow the few-shot learning protocol where only a limited number of training data are used.**

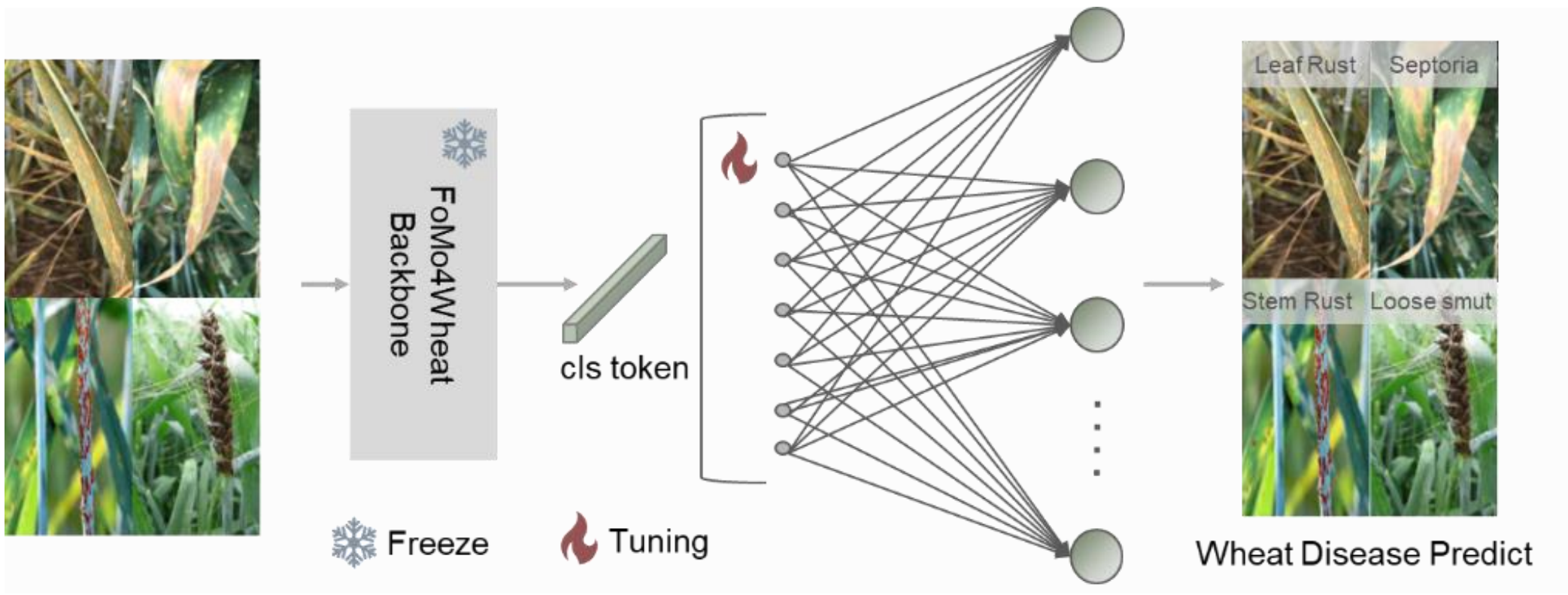

**Supplementary Figure 1.** Classification task head.

***Wheat Growth Stage Classification.*** **The first task is wheat growth stage detection where the model is tasked to predict the correct growth stage. We adopt the WGSP dataset**[2] **for wheat growth stage classification. The dataset contains 3353 RGB images of wheat canopies collected from field trials conducted in China, the United Kingdom, and the United States between 2018 and 2021. Spanning four growing seasons, the dataset covers 263 wheat varieties and encompasses six critical growth stages: tillering, jointing, booting,** Anthesis**, grain filling, and** maturity**. The dataset reflects a wide range of geographic locations, climatic conditions, planting densities, and canopy color and structural characteristics across different wheat varieties. We partition it into 671 training images and 2,682 testing images. Under a reduced training data protocol, that is, 75%, 50%, and 25% data are sampled from the original training set, we performed five independent random sampling per data fraction to mitigate stochastic sampling bias.**

***Wheat Disease Classification*. The second task is wheat disease classification. We collected images from five public datasets of CerealConv**[3]**, WFD**[4]**, AWDD**[5]**, MSWDD**[6]**, and LWDCD**[7]**, to construct a comprehensive dataset of healthy wheat and other eight disease types including Brownrust, Mildew, Septoria, Yellowrust, Stemrust, Healthy, Wheatscab, and Yellowdwarf, with 4000 images in total. These images were obtained from multiple countries in Europe, North America, and Asia between 2019 and 2024, reflecting the sensitivity and resistance of different wheat varieties to diseases and covering images of different varieties, periods, and environmental conditions. Based on this dataset, we established the Wheat Disease Classification benchmark, partitioned into 320 training samples and 3,680 testing samples. The identical random five-fold sampling trials were conducted to ensure statistical soundness.**

***Evaluation Metrics*. We report the balanced accuracy and mean average precision as evaluation metrics.**

- **Balanced Accuracy (BA): Balanced Accuracy calculates the average recall across all classes. For $k$ classes, let $TP_i$ and $FN_i$ denote the true positives and false negatives of the $i$-th class, respectively. BA is defined by**

$$BA = \frac{1}{k}\sum_{i=1}^{k}\frac{TP_i}{TP_i + FN_i}.$$

- **Mean Average Precision (mAP): $AP$ measures the precision-recall trade-off of a single class. For class $i$, let $p_i(r)$ denote the precision at the recall level $r$. $AP$ is the area under the precision-recall curve, which takes the form**

$$AP_i = \int_0^1 p_i(r)\, dr .$$

**$mAP$ generalizes $AP$ to multi-class scenarios by computing the mean of AP across all classes. For $k$ classes, $mAP$ is defined by**

$$mAP = \frac{1}{k}\sum_{i=1}^{k} AP_i .$$

## 5. Detection Tasks

***Detection Task Head***. Here **we evaluate the adaptability of the backbone on the detection tasks where a model should output bounding boxes to indicate what objects are and where they are. We adopt the Mask R-CNN head[8] for detection, and the backbone parameters were frozen. The constructed baseline is identical the ViT-Adapter[9]. According to Supplementary Figure 2, the standard Mask R-CNN head includes a Feature Pyramid Network (FPN) used to generate multi-scale features, a Region Proposal Network (RPN) head used to generate Region-of-Interest (RoI) object proposals, a fully-connected branch for bounding box prediction, and a fully-convolutional branch for mask prediction. For detection evaluation, we neglect the mask prediction head.**

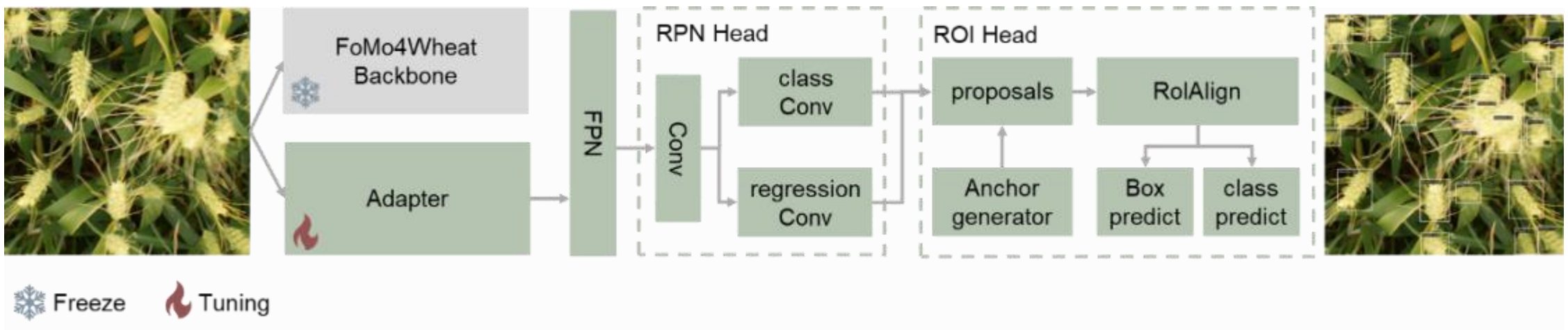

**Supplementary Figure 2.** Detection Task Head.

***Wheat Head Detection From Ground-based Imagery*. We use the Global Wheat Head Detection (GWHD) Dataset[10] to evaluate wheat head detection. The dataset is developed for an international competition, with 4,700 high-resolution RGB images from various locations around the world between 2016 and 2019, containing 190,000 labeled wheat heads. The images cover multiple genotypes at different growth stages and include data from several countries, including Europe, North America, Australia, and Asia, under various planting densities and environmental conditions. The dataset is divided into 3,607 training images, 1,448 validation images, and 1,382 testing images.**

***Wheat Spike Detection From UAV Imagery*. From 2021 to 2024, our team conducted field experiments in various locations including Yangling, Nanchong, Baima, and Jurong, China, and developed a drone-based wheat spike detection dataset. Visible light images were captured at heights of both six and ten meters, with a resolution of 4608x3456 pixels. The high-resolution images were then cropped to 224x224 pixels to highlight wheat spikes. To assess the cross-domain generalization of ground-based wheat head detection models, we established a UAV-based testing set comprising images acquired at different altitudes: 100 images at GSD 0.6mm elevation (Nanchong) and 120 images at GSD 1.2mm elevation (Yangling).**

***Evaluation Metrics*. We use the conventional average precision (AP) to score detection performance, defined as follows.**

- **Average Precision (AP)**: **Given the predicted and ground-truth bounding boxes, the intersection over union (IoU) metric is first computed by measuring the intersection area over union area between the two boxes as**

$$IoU = \frac{\text{Area of Intersection}}{\text{Area of Union}}.$$

**Different from the class $AP$ definition above, detection $AP$ measures the precision-recall trade-off across multiple IoU thresholds. For a single object class, let $p(r)$ denote the precision at recall level $r$. $AP$ is defined by**

$$AP = \int_0^1 p(r)\ dr\ .$$

- **Average Precision at 50% IoU (AP50)**: Further, we highlight **the** $AP50$ **metric where the IoU threshold is set to 0.5, relaxing the localization accuracy to focus on classification performance, defined by**

$$AP50 = \int_0^1 p_{0.5}(r)\ dr$$

## 6. Counting Tasks

***Counting Task Head*. Different from detection, counting does not need to predict where objects are, but how many they are. We adopt the Point quEry Transformer (PET)[11] as our baseline.** The PET model employs a point-based regression paradigm to solve dense object counting. PET redefines dense counting as a decomposable point-query process, where a dynamic quadtree splitting mechanism is used to adaptively adjust querying point density, enabling coverage of both sparse and dense regions. The model integrates CNN features with positional encoding to jointly represent query points and employs a progressive rectangular window attention mechanism to hierarchically capture both global context and local details, balancing perspective priors and computational efficiency. **We follow this model and incorporate its counting head for point prediction. Note that the backbone parameters were frozen. We evaluate on two counting tasks of wheat leaf counting and rice leaf counting.**

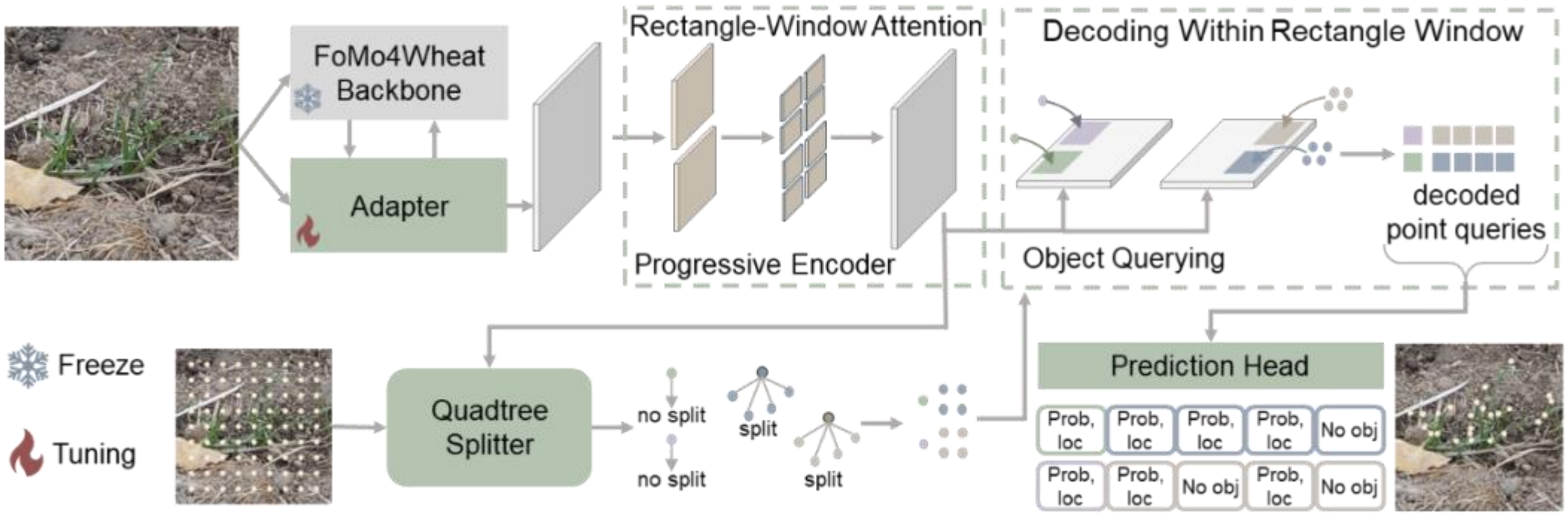

**Supplementary Figure 3.** Counting Task Head.

***Wheat Leaf Counting*. We first evaluate on** wheat leaf counting where a model is required to count all leaf tips. Between 2020 and 2022[12], our team conducted field experiments in Xuzhou, Baima, Henan, and Jurong, China, and created a leaf counting benchmark using our self-developed PhenoArm imaging equipment and outdoor monitoring devices with the Hunting camera. Image data on the canopy structure at different growth stages before wheat tillering were captured from 0° and 45° angles. Each leaf tip is labeled with a dot. Five experimenters were involved in data labelling and were cross-checked with each other for correctness. **1,508 images were used for training, and 379 images are used for testing.**

***Rice Leaf Counting*. To justify cross-species generalization of FoMo4Wheat, we also evaluate the counting performance on rice crop.** Between 2020 and 2022, our team conducted field experiments in Nanjing, Sanya, Danyang, covering over 1,000 rice varieties. Using drones, we collected canopy structure data at different growth stages of rice prior to tillering. We call this dataset as the Rice Leaf Counting dataset, partitioned into 1,360 training images and 340 testing images.

***Evaluation Metrics*. We follow the standard error metrics to assess the counting performance, including:**

- **Coefficient of Determination ($R^2$): $R^2$ evaluates the proportion of variance in the ground-truth counts that is predictable from the predicted counts. For $N$ samples, let $y_i$ and $\hat{y}_i$ denote the true and predicted counts of the $i$-th sample, respectively, and let $\bar{y}$** be the mean of all true counts. $R^2$ is defined by

$$R^2 = 1 - \frac{\sum_{i=1}^{n}(y_i - \hat{y}_i)^2}{\sum_{i=1}^{n}(y_i - \bar{y})^2}$$

- **Mean Absolute Error (MAE): MAE measures the average absolute deviation between predicted and ground-truth counts. For $N$ samples, let $y_i$ and $\hat{y}_i$ denote the true and predicted counts of the $i$-th sample. MAE is defined by**

$$MAE = \frac{1}{N}\sum_{i=1}^{N}|y_i - \hat{y}_i| .$$

- **Root Mean Squared Error (RMSE)**: **RMSE measures the average squared deviation, revealing the counting robustness. For $N$ samples, RMSE takes the form**

$$RMSE = \frac{1}{N}\sum_{i=1}^{N}\sqrt{(y_i - \hat{y}_i)^2} .$$

## 7. Segmentation Tasks

***Segmentation Task Head*. Segmentation requires per-pixel label prediction.** We adopt the Mask2Former head[13] for segmentation (Supplementary Figure 4). The Mask2Former head includes a pixel decoder used to generate multi-scale features and a transformer decoder used to predict object class and object mask. The mask head and the adapter were trained, while the backbone parameters were frozen during training. We evaluate on four different segmentation tasks.

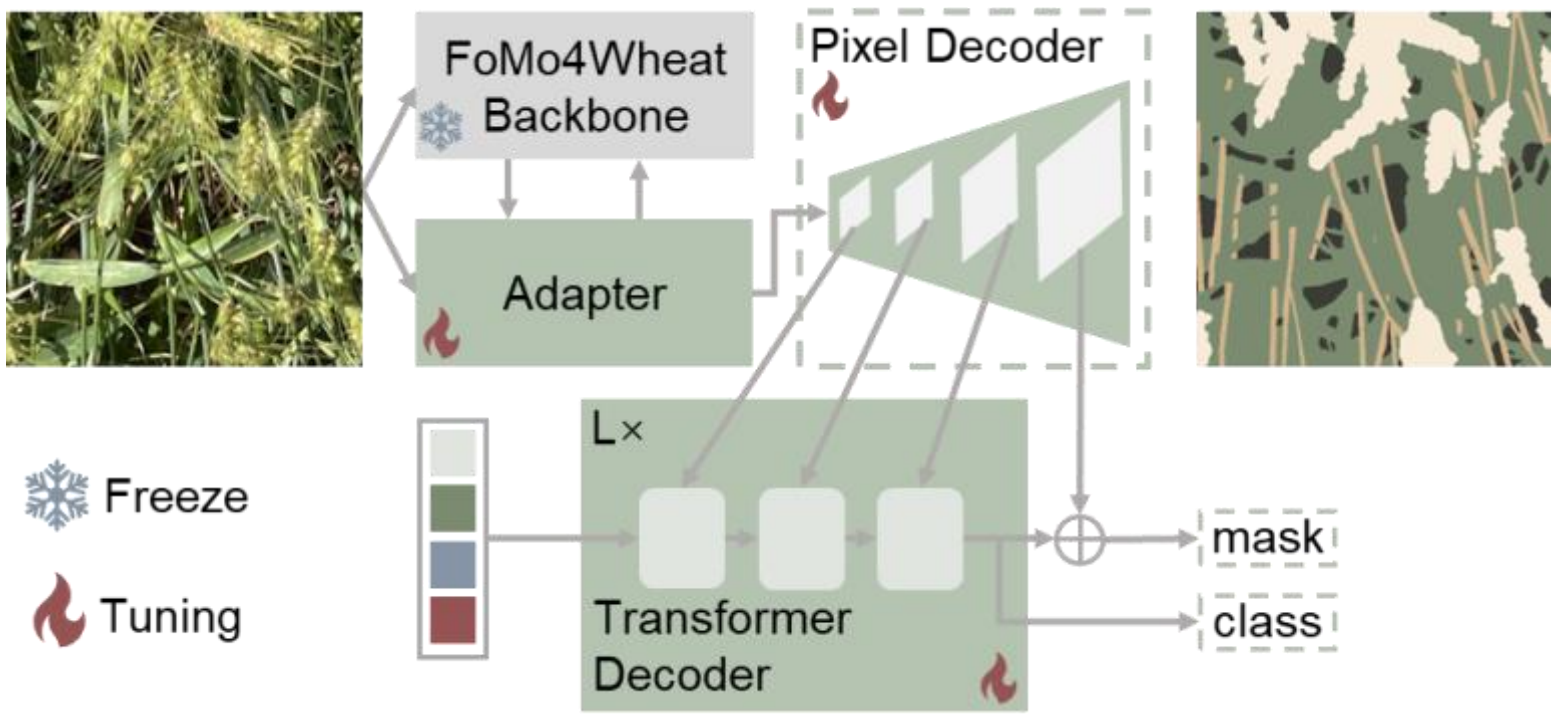

**Supplementary Figure 4.** Segmentation Task Head.

***Wheat Organ Segmentation*.** The first is wheat organ segmentation. The Global Wheat Full Semantic Segmentation (GWFSS)[14] dataset is employed. It is a recently released international competition dataset that includes 1,096 high-resolution RGB images collected from around the world between 2020 and 2024. The dataset contains four categories for wheat organ segmentation. The images cover a variety of genotypes at different growth stages and include data from collaborative institutions across Switzerland, Belgium, UK, China , Mexico, Australia, France, Morocco, Japan, Canada collected under various planting densities and environmental conditions. The dataset is partitioned into 876 training images and 220 testing images.

***Rice Organ Segmentation*. The second is multi-class rice segmentation. We validate on another crop variety to justify the cross-species generalization of the backbone.** Our team developed the RiceSEG[15] image segmentation dataset through international collaboration, focusing on rice fields with weeds and duckweed backgrounds. The images were collected between 2012 and 2023 by 10 institutions from 10 locations across five countries, including China, Japan, India, the Philippines, and Tanzania, covering over 1,000 rice varieties. The images were captured using various camera types, such as digital SLRs, portable action cameras, and smartphones. The cameras were positioned 1-2 meters above the canopy and oriented toward the canopy in different directions (0°-90°). The dataset has 2,462 training images and 616 testing images.

***Multi-Crop Segmentation*. The third is crop segmentation. We use** the VegAnn[16] dataset, which is a binary segmentation dataset designed to differentiate vegetation, including both healthy and senescent plant parts, from background elements such as soil and crop residues. It comprises 3,775 high-resolution RGB images collected from diverse regions using various acquisition systems and platforms. The dataset encompasses 9 crop types, spanning different growth stages, climatic conditions, and soil types, and includes data from multiple countries such as France, China, Japan, Belgium, Australia, and others. The dataset is partitioned into 3,020 training images and 755 test images.

***Crop and Weed Segmentation*. The forth task is to segment crop and weed.** The Crop and Weed dataset[17] is adopted, which is a large-scale, multi-modal dataset designed for crop and weed segmentation. The data was collected from multiple commercial agricultural sites and specially cultivated in outdoor experimental plots in Austria, spanning various growing seasons between 2018 and 2021. It covers diverse lighting conditions, weather scenarios, soil types, and combinations of crops and weeds, while also including negative samples with no visible vegetation and backgrounds containing debris. The dataset contains 7,705 images and approximately 112,000 annotated instances. We adopt the Fine24 subvariant of the Crop and Weed dataset as the benchmark (16 crop and 8 weed categories), partitioned into 6,164 training images and 1,541 test images.

***Evaluation Metrics*. We report two segmentation metrics:**

- **Mean Intersection over Union (mIoU): $mIoU$ measures average region overlap between predicted and ground-truth masks across all classes. For $C$ classes, let $TP_i$, $FP_i$, and $FN_i$ denote the true positives, false positives, and false negatives for of the $i$-th class.** mIoU **is defined as**

$$mIoU = \frac{1}{C}\sum_{i=1}^{C}\frac{TP_i}{TP_i + FP_i + FN_i}.$$

- **Mean Accuracy (mACC)**: Different from $mIoU$, $mAcc$ **calculates average pixel-wise classification accuracy across all classes. Condiering $C$ classes,** $mAcc$ takes the form

$$mAcc = \frac{1}{C}\sum_{i=1}^{C}\frac{TP_i + TN_i}{TP_i + TN_i + FP_i + FN_i}$$

## Supplementary Tables

**Supplementary Table 5.** The compostion of the ImAgWheat Dataset. (Note: Crop_Number: The number of images after data preprocessing; Mix Pixels: The public dataset covers mixed resolution images)

| Country | Site | Lat | Long | Year | No. of genotypes | Device | Image size | Number | Crop_Number |
|---|---|---|---|---|---|---|---|---|---|
| China | Baima | 32°3’41.6”N | 118°47’29.6”E | 2021 | 120 | Sony RX0 | 4800*2704 | 10453 | 83624 |
| | Baima | 32°3’41.6”N | 118°47’29.6”E | 2021 | 120 | Sony RX0 | 4800*3200 | 4920 | 59040 |
| | Jurong | 31°57'00.00"N | 119°10'00.00"E | 2021 | 5 | Sony RX0 | 4800*3200 | 600 | 7200 |
| | Xuzhou | 34°12’18.3”N | 117°16’41.9”E | 2021 | 5 | Sony RX0 | 4800*3200 | 736 | 8832 |
| | Xinxiang | 35°18’7.6”N | 113°55’12.7”E | 2021 | 120 | Sony RX0 | 4800*3200 | 14291 | 171492 |
| | Baima | 32°3’41.6”N | 118°47’29.6”E | 2021 | 120 | Canon EOS X5 | 1024*1024 | 1386 | 1386 |
| | Xuzhou | 34°12’18.3”N | 117°16’41.9”E | 2021 | 5 | Canon EOS X5 | 1024*1024 | 688 | 688 |
| | Jurong | 31°57'00.00"N | 119°10'00.00"E | 2021 | 5 | Canon EOS X5 | 1024*1024 | 309 | 309 |
| | Xinjiang | 44°18'8.28'' N | 86°2'12.984'' E | 2021 | 120 | Canon EOS X5 | 4800*3200 | 4773 | 57276 |
| | Jinan | 36°39’58.7”N | 117°4’12.8”E | 2022 | 565 | Huniting camera 800M | 3264*2448 | 57446 | 344676 |
| | Xinxiang | 35°18’7.6”N | 113°55’12.7”E | 2023 | 360 | Huniting camera 800M | 3264*2448 | 57511 | 345066 |
| | Zhoukou | 33°37’50.3”N | 114°38’32.5”E | 2023 | 565 | Huniting camera 800M | 3264*2448 | 32109 | 192654 |
| | Baima | 32°3’41.6”N | 118°47’29.6”E | 2024 | 565 | Canon EOS X5 | 4800*3200 | 13254 | 159048 |
| | Nanyang | 33°0’3.8”N | 112°31’45.4”E | 2024 | 565 | Canon EOS X5 | 4800*3200 | 6101 | 73212 |
| | Yangling | 34°16’21.8”N | 108°4’52.1”E | 2022 | 565 | Sony RX0 | 4800*2704 | 27047 | 216376 |
| | Yangling | 34°16’21.8”N | 108°4’52.1”E | 2022 | 565 | Sony RX0 | 1200*800 | 27046 | 54092 |
| | Yangling | 34°16’21.8”N | 108°4’52.1”E | 2023 | 565 | Sony RX0 | 4800*2704 | 36078 | 288624 |
| | Yangling | 34°16’21.8”N | 108°4’52.1”E | 2023 | 565 | Sony RX0 | 1200*800 | 36077 | 72154 |
| | Yangling | 34°16’21.8”N | 108°4’52.1”E | 2024 | 565 | PhenoArm | 4800*3200 | 32710 | 392520 |
| France | Gréoux | 43°45’30.0”N | 5°53’10.0”E | 2016 | 19 | Canon EOS X5 | 1083 * 719 | 18203 | 91015 |
| | Gréoux | 43°45’30.0”N | 5°53’10.0”E | 2017 | 20 | Canon EOS X5 | 1083 * 719 | 18040 | 90200 |
| | Gréoux | 43°45’30.0”N | 5°53’10.0”E | 2018 | 20 | Canon EOS X5 | 1083 * 719 | 20162 | 100810 |
| | Muizon | 49°16’30.0”N | 3°53’10.0”E | 2019 | / | Canon EOS X5 | 1024 * 768 | 177 | 885 |
| | Boigneville | 48°20’10.0”N | 2°22’15.0”E | 2019 | / | Canon EOS X5 | 1024 * 768 | 1009 | 5045 |
| | Chalons | 48°57’20.0”N | 4°21’30.0”E | 2019 | / | Canon EOS X5 | 1024 * 768 | 312 | 1560 |
| | St-Hilaire | 46°45’10.0”N | 1°53’30.0”E | 2019 | / | Canon EOS X5 | 1024 * 768 | 253 | 1265 |
| | Nimes | 43°50’15.0”N | 4°21’40.0”E | 2019 | / | Canon EOS X5 | 1024 * 768 | 205 | 1025 |
| | Toulouse | 43°36’15.0”N | 1°26’40.0”E | 2021 | / | Sigma SD14 | 1024 * 1024 | 115 | 575 |
| | Avignon | 43°56’30.0”N | 4°48’20.0”E | 2021 | / | Sigma SD14 | 4068 * 3072 | 270 | 2160 |
| | Paris | 48°51’30.0”N | 2°21’15.0”E | 2021 | / | NIKOND 5200 | 6000 * 4000 | 200 | 2400 |
| Australia | Gatton Research Station | 27°33’00.0”S | 152°19’48.0”E | 2015 | / | Canon EOS 5500 | 5184*3456 | 4939 | 59268 |
| | Gatton Research Station | 27°33’00.0”S | 152°19’48.0”E | 2016 | / | Canon EOS 5500 | 5184*3456 | 4221 | 50652 |
| | Gatton Darbalara Farm | 27°38’14.6”S | 152°23’34.1”E | 2021 | / | GECKOCAM | 1024 * 1024 | 169 | 845 |
| | Gatton Darbalara Farm | 27°38’14.6”S | 152°23’34.1”E | 2022 | / | GECKOCAM | 1200 * 800 | 1742 | 8710 |
| | Gatton Darbalara Farm | 27°38’14.6”S | 152°23’34.1”E | 2022 | / | LITERAL | 4800*3200 | 5843 | 70116 |
| Japan | Tokyo | 35°41’30.0”N | 139°41’15.0”E | 2022 | / | Canon EOS X5 | 1280 * 1040 | 1485 | 7425 |
| Switzerland | Eschikon Field Station | 47°26’30.0”N | 8°41’15.0”E | 2018 | / | FIP | 2817*1876 | 2751 | 33012 |
| America | Nebraska--Lincoln | 40°49’30.0”N | 96°41’15.0”W | 2018 | / | iPad Pro | 4032 * 3024 | 476 | 14280 |
| Ethiopia[18] | Holeta wheat farm | 9°02’60.00”N | 38°29’59.99”E | 2021 | / | EOS 5D Mark III | Mix Pixels | 407 | 407 |
| UK and Ireland[3] | Norwich Research Park | 52°37’18.0”N | 1°13’30.0”E | 2019 | / | Smartphone | Mix Pixels | 19159 | 19159 |
| India[19] | Punjab | 30°58’00.0”N | 74°37’00.0”E | 2020 | / | Custom Android application | Mix Pixels | 11397 | 11397 |
| | Haryana | 28°27’33.0”N | 77°01’23.0”E | 2020 | / | Custom Android application | Mix Pixels | | |

**Supplementary Table 6.** Model details of the ViT-B/L/G networks used in this work.

| Model | Initialization | Input Resolution | Architect ure | Drop-rate | Learning Rate | Batch Size | Embed Dim | Hea ds | Bloc ks | FFN Layer |
|---|---|---|---|---|---|---|---|---|---|---|
| Base | Distilled | 224 | ViT-B/14 | 0 | 1.00E-04 | 3072 | 768 | 12 | 18 | MLP |
| Base | Distilled | 518 | ViT-B/14 | 0 | 5.00E-05 | 768 | 768 | 12 | 18 | MLP |
| Large | Distilled | 224 | ViT-L/14 | 0 | 1.00E-04 | 3072 | 1024 | 16 | 24 | MLP |
| Large | Distilled | 518 | ViT-L/14 | 0 | 1.00E-04 | 768 | 1024 | 16 | 24 | MLP |
| Base | Scratch | 224 | ViT-B/14 | 0.4 | 1.00E-04 | 3072 | 768 | 12 | 18 | SwiGLU |
| Base | Scratch | 518 | ViT-B/14 | 0.4 | 5.00E-05 | 768 | 768 | 12 | 18 | SwiGLU |
| Large | Scratch | 224 | ViT-L/14 | 0.4 | 1.00E-04 | 3072 | 1024 | 16 | 24 | SwiGLU |
| Large | Scratch | 518 | ViT-L/14 | 0.4 | 1.00E-04 | 768 | 1024 | 16 | 24 | SwiGLU |
| Giant | Pre-trained | 224 | ViT-g/14 | 0.4 | 1.00E-04 | 1536 | 1536 | 24 | 40 | SwiGLU |
| Giant | Pre-trained | 518 | ViT-g/14 | 0.4 | 2.00E-05 | 384 | 1536 | 24 | 40 | SwiGLU |
| Giant | Random | 224 | ViT-g/14 | 0.4 | 1.00E-04 | 1536 | 1536 | 24 | 40 | SwiGLU |
| Giant | Random | 518 | ViT-g/14 | 0.4 | 2.00E-05 | 384 | 1536 | 24 | 40 | SwiGLU |

**Supplementary Table 7.** Statistics of downstream task dataset.

| Task | Domain | Class | Label Type | Label Classes | Image Num | Training | Test |
|---|---|---|---|---|---|---|---|
| Classification | Wheat Growth Stage | 6 | Category | Tillering, Jointing, Booting, Anthesis, Grain Filling, Maturity | 3353 | 671 | 2682 |
| | Wheat Disease | 8 | Category | Brownrust, Mildew, Septoria, Yellowrust, Stemrust, Wheatscab, Yellowdwarf, Healthy, | 4000 | 320 | 3680 |
| Detection | Wheat Spikes | 1 | Box | Spike | 6437 | 3607 | 2830 |
| | UAV Wheat Spikes (6m) | 1 | Box | Spike | 100 | / | 100 |
| | UAV Wheat Spikes (10m) | 1 | Box | Spike | 120 | / | 120 |
| Counting | Wheat Leaf | 1 | Point | Leaf | 1887 | 1508 | 379 |
| | Rice Leaf | 1 | Point | Leaf | 1700 | 1360 | 340 |
| Segmentation | Wheat Organ | 4 | Mask | Background, Head, Stem, Leaf | 1096 | 876 | 220 |
| | Rice Organ | 6 | Mask | Background, Green, Senescent, Panicle, Weed, Duckweed | 3078 | 2462 | 216 |
| | Multi-Crop | 9 | Mask | Background, Wheat, Rapeseed, Maize, Sunflower, Sugarbeet, Mix, Rice, Potato | 3267 | 2614 | 653 |
| | Crop and Weed | 25 | Mask | Background, Maize, Sugar beet, Soy, Sunflower, Potato, Pea, Bean, Pumpkin, Grasses, Amaranth, Goosefoot, Knotweed, Corn spurry, Chickweed, Solanales, Potato weed, Chamomile, Thistle, Mercuries, Geranium, Crucifer, Poppy, Plantago, Labiate | 7705 | 6164 | 1541 |

**Supplementary Table 8.** Classification results on the wheat growth Stage and Disease. (Note: We conducted experiments using 100%, 75%, 50%, and 25% of the full training dataset)

| Method | Architecture | Train Dataset | Growth Stage | | Disease | |
|---|---|---|---|---|---|---|
| | | | BA | MAP | BA | MAP |
| SOTA | Giant | 100% | 91.3 | 95.6 | 87.6 | 92.9 |
| FoMo4Wheat | | | **93.0** | **96.8** | **92.6** | **96.5** |
| SOTA | Large | | 90.4 | 95.7 | 87.7 | 93.1 |
| FoMo4Wheat | | | **91.5** | **96.1** | **92.3** | **96.7** |
| SOTA | Base | | 91.0 | 95.6 | 87.8 | 92.9 |
| FoMo4Wheat | | | **92.1** | **96.3** | **90.1** | **94.6** |
| SOTA | Giant | 75% | 90.3 (±0.33) | 94.6 (±0.59) | 87.4 (±0.72) | 93.1 (±0.61) |
| FoMo4Wheat | | | **93.1** (±0.32) | **96.5** (±0.30) | **92** (±0.68) | **96.4** (±0.38) |
| SOTA | Large | | 90.0 (±0.32) | 95.2 (±0.25) | 86.6 (±0.86) | 92.3 (±0.48) |
| FoMo4Wheat | | | **91.4** (±0.51) | **95.6** (±0.08) | **90.3** (±0.98) | **95.6** (±0.61) |
| SOTA | Base | | 90.1 (±0.65) | 94.9 (±0.51) | 85.8 (±0.88) | 91.7 (±0.69) |
| FoMo4Wheat | | | **92.1** (±0.43) | **95.9** (±0.30) | **87.4** (±0.48) | **92.7** (±0.63) |
| SOTA | Giant | 50% | 89.4 (±1.03) | 94.0 (±0.57) | 84.6 (±1.42) | 90.6 (±1.11) |
| FoMo4Wheat | | | **92.4** (±0.58) | **95.3** (±1.70) | **90.1** (±0.57) | **95.1** (±0.40) |
| SOTA | Large | | 88.7 (±0.99) | 94.1 (±0.59) | 84.7 (±1.23) | 90.1 (±1.01) |
| FoMo4Wheat | | | **90.8** (±0.48) | **94.9** (±0.53) | **88.1** (±0.74) | **94.1** (±0.55) |
| SOTA | Base | | 89.4 (±0.79) | 94.2 (±0.37) | 82.9 (±0.94) | 89 (±0.79) |
| FoMo4Wheat | | | **91.1** (±0.49) | **95.3** (±0.71) | **85.5** (±0.94) | **90.8** (±0.57) |
| SOTA | Giant | 25% | 86.7 (±1.07) | 91.4 (±1.49) | 73.6 (±2.09) | 78.8 (±2.45) |
| FoMo4Wheat | | | **90.8** (±0.90) | **95.1** (±0.90) | **82.6** (±2.15) | **88.9** (±1.30) |
| SOTA | Large | | 86.0 (±1.26) | 91.1 (±1.47) | 74.6 (±2.42) | 80.6 (±2.71) |
| FoMo4Wheat | | | **88.6** (±1.01) | **93.3** (±1.27) | **80.6** (±2.13) | **88.0** (±1.86) |
| SOTA | Base | | 86.5 (±1.46) | 91.5 (±1.61) | 72.3 (±1.70) | 78.9 (±1.37) |
| FoMo4Wheat | | | **90.2** (±0.74) | **94.5** (±0.34) | **78.5** (±1.69) | **84.9** (±1.25) |

**Supplementary Table 9.** Detection results on the ground-based and UAV-based wheat spikes. Best performance is in **boldface**.

| Method | Architecture | Ground-based Wheat Spike (GSD 0.4mm) | | UAV Wheat Spike | | | |
|---|---|---|---|---|---|---|---|
| | | | | GSD 0.6mm | | GSD 1.2mm | |
| | | AP | AP-50 | AP | AP-50 | AP | AP-50 |
| SOTA | Giant | 36.0 | 73.4 | 34.2 | 78.2 | 27.4 | 55.4 |
| FoMo4Wheat | | **36.6** | **74.0** | **35.4** | **80.2** | **28.3** | **56.0** |
| SOTA | Large | 33.7 | 70.4 | 28.4 | 70.4 | 22.9 | 48.0 |
| FoMo4Wheat | | 34.6 | 72.0 | 32.1 | 75.9 | 25.3 | 50.8 |
| SOTA | Base | 33.0 | 69.5 | 27.1 | 69.9 | 23.7 | 48.5 |
| FoMo4Wheat | | 33.3 | 70.0 | 28.6 | 71.9 | 24.4 | 50.4 |
| GWHD | / | 34.8 | / | / | / | / | / |

**Supplementary Table 10.** Counting results on the Wheat Leaf dataset and UAV Rice Leaf dataset. Best performance is in **boldface**.

| Method | Architecture | Wheat Leaf | | | Rice Leaf | | |
|---|---|---|---|---|---|---|---|
| | | MAE | MSE | R2 | MAE | MSE | R2 |
| SOTA | Giant | 9.51 | 14.26 | 0.88 | 11.16 | 15.17 | 0.90 |
| FoMo4Wheat | | **9.19** | **14.11** | **0.89** | **10.77** | **14.72** | **0.91** |
| SOTA | Large | 10.24 | 15.46 | **0.86** | 11.68 | 15.34 | 0.90 |
| FoMo4Wheat | | **10.15** | **15.40** | **0.86** | **11.06** | **14.69** | **0.91** |
| SOTA | Base | 10.34 | 16.19 | 0.85 | 12.39 | 16.89 | 0.88 |
| FoMo4Wheat | | **9.85** | **14.77** | **0.88** | **11.59** | **15.49** | **0.90** |

**Supplementary Table 11.** Segmentation results on the GWFSS dataset for wheat organ segmentation. Best performance is in **boldface**.

| Method | Architecture | Class IOU | | | | mIoU |
|---|---|---|---|---|---|---|
| | | background | head | stem | leaf | |
| SOTA | Giant | 84.80 | 79.70 | 37.48 | 80.28 | 70.57 |
| FoMo4Wheat | | **85.97** | **84.48** | **52.73** | **82.76** | **76.49** |
| SOTA | Large | 84.58 | 79.10 | 36.71 | 80.00 | 70.10 |
| FoMo4Wheat | | **86.01** | **83.72** | **49.90** | **82.30** | **75.48** |
| SOTA | Base | 84.39 | 77.92 | 34.51 | 79.44 | 69.07 |
| FoMo4Wheat | | **85.90** | **83.31** | **47.67** | **82.21** | **74.77** |
| GWFSS | / | 84.51 | 82.85 | 44.92 | **82.35** | 73.66 |
| **Method** | **Architecture** | **Class Acc** | | | | **mAcc** |
| | | background | head | stem | leaf | |
| SOTA | Giant | 90.63 | 88.43 | 46.13 | 91.23 | 79.10 |
| FoMo4Wheat | | **91.80** | **92.80** | **67.74** | **91.11** | **85.86** |
| SOTA | Large | 90.40 | 87.81 | 45.04 | 91.24 | 78.62 |
| FoMo4Wheat | | **92.06** | **92.59** | **64.47** | **90.66** | **84.94** |
| SOTA | Base | 90.27 | 86.93 | 42.61 | 90.98 | 77.70 |
| FoMo4Wheat | | **91.15** | **92.16** | **60.14** | **91.68** | **83.78** |

**Supplementary Table 12.** Segmentation results on the RiceSEG dataset. Best performance is in **boldface**.

| Method | Architecture | Class IOU | | | | | | mIoU |
|---|---|---|---|---|---|---|---|---|
| | | background | green | senescent | panicle | weed | duckweed | |
| SOTA | Giant | 89.40 | 86.16 | 50.16 | 67.45 | 62.65 | 64.01 | 69.97 |
| FoMo4Wheat | | **89.48** | **86.71** | **53.28** | **74.69** | **73.32** | **69.77** | **74.54** |
| SOTA | Large | 89.32 | 86.11 | 49.41 | 66.71 | 63.76 | 63.94 | 69.87 |
| FoMo4Wheat | | **89.37** | **86.60** | **52.72** | **73.34** | **69.79** | **67.49** | **73.22** |
| SOTA | Base | 89.24 | 85.60 | 49.88 | 65.42 | 57.74 | 62.32 | 68.37 |
| FoMo4Wheat | | **89.29** | **86.44** | **51.77** | **72.35** | **68.34** | **67.06** | **72.54** |
| Method | Architecture | Class Acc | | | | | | mAcc |

| | | background | green | senescent | panicle | weed | duckweed | |
|---|---|---|---|---|---|---|---|---|
| SOTA | Giant | 94.78 | 92.91 | 62.03 | 80.81 | 72.05 | 74.45 | 79.50 |
| FoMo4Wheat | | **94.71** | **93.10** | **65.04** | **86.72** | **82.37** | **78.23** | **83.36** |
| SOTA | Large | **94.75** | 92.84 | 60.04 | 80.92 | 74.45 | 75.84 | 79.81 |
| FoMo4Wheat | | 94.64 | **92.99** | **65.44** | **86.48** | **78.47** | **77.55** | **82.59** |
| SOTA | Base | **94.58** | 92.73 | 61.25 | 80.01 | 68.65 | 72.96 | 80.29 |
| FoMo4Wheat | | 94.55 | **93.00** | **64.53** | **85.85** | **77.25** | **76.27** | **81.91** |

**Supplementary Table 13.** Segmentation results on the VegAnn dataset for multi-crop segmentation. Best performance is in **boldface**.

| Method | Architecture | Class IOU | | | | | | | | | mIoU |
|---|---|---|---|---|---|---|---|---|---|---|---|
| | | background | Wheat | Rapeseed | Maize | Sunflower | Sugarbeet | Mix | Rice | Potato | |
| SOTA | Giant | 85.91 | 77.86 | 80.95 | 66.63 | 83.22 | 86.80 | 64.60 | 85.60 | 73.68 | 78.36 |
| FoMo4Wheat | | **87.45** | **86.66** | **94.40** | **92.30** | **94.82** | **96.22** | **78.20** | **91.30** | **95.03** | **90.71** |
| SOTA | Large | 85.44 | 75.89 | 80.77 | 71.81 | 65.97 | 82.40 | 63.20 | 83.60 | 64.44 | 74.84 |
| FoMo4Wheat | | **87.95** | **87.39** | **94.47** | **90.21** | **85.25** | **81.70** | **79.60** | **89.30** | **88.73** | **87.18** |
| SOTA | Base | 84.25 | 67.72 | 63.28 | 56.11 | 58.34 | 50.66 | 59.60 | 86.10 | 53.33 | 64.37 |
| FoMo4Wheat | | **87.83** | **85.27** | **94.45** | **90.18** | **85.07** | **76.87** | **76.10** | **89.10** | **82.75** | **85.28** |

| Method | Architecture | Class Acc | | | | | | | | | mAcc |
|---|---|---|---|---|---|---|---|---|---|---|---|
| | | background | Wheat | Rapeseed | Maize | Sunflower | Sugarbeet | Mix | Rice | Potato | |
| SOTA | Giant | 93.21 | 85.68 | 88.82 | 70.08 | 91.84 | 88.14 | 85.10 | 90.00 | 84.93 | 86.42 |
| FoMo4Wheat | | **93.30** | **92.98** | **97.26** | **95.93** | **97.20** | **97.63** | **88.10** | **94.60** | **97.54** | **94.95** |
| SOTA | Large | 93.09 | 84.06 | 87.84 | 76.98 | 74.69 | 83.52 | 85.50 | 89.90 | 78.37 | 84.18 |
| FoMo4Wheat | | **93.54** | **94.40** | **96.72** | **92.93** | **90.94** | **88.61** | **88.60** | **92.40** | **97.96** | **92.90** |
| SOTA | Base | **93.24** | 75.70 | 80.76 | 60.41 | 61.17 | 57.75 | 87.50 | 90.60 | 79.60 | 77.10 |
| FoMo4Wheat | | 93.18 | **92.57** | **97.31** | **93.64** | **91.12** | **88.80** | **87.90** | **92.20** | **91.65** | **92.04** |

**Supplementary Table 14.** Segmentation results on the Crop and Weed dataset for crop and weed segmentation. Best performance is in **boldface**.

| Method | | Vit-Adapter | FoMo4Wheat | Vit-Adapter | FoMo4Wheat | Vit-Adapter | FoMo4Wheat |
|---|---|---|---|---|---|---|---|
| Architecture | | Giant | | Large | | Base | |
| Class IOU | Background | 99.46 | **99.49** | 99.46 | **99.48** | 99.44 | **99.46** |
| | Maize | 87.35 | **89.44** | **86.87** | 88.64 | 86.36 | **87.84** |
| | Sugar beet | 81.97 | **88.36** | 86.37 | **89.18** | 75.10 | **87.60** |
| | Soy | 75.32 | **82.57** | 76.20 | **82.22** | 70.89 | **78.12** |
| | Sunflower | **90.56** | 90.42 | 87.63 | **92.71** | 89.97 | **91.20** |
| | Potato | 79.32 | **86.75** | 79.93 | **86.95** | 76.29 | **91.78** |
| | Pea | 75.70 | **77.59** | 75.32 | **75.90** | 75.13 | **75.67** |
| | Bean | 79.95 | **82.61** | 70.37 | **85.08** | 72.80 | **81.03** |
| | Pumpkin | 89.03 | **90.46** | **91.96** | 91.63 | 87.31 | **91.41** |
| | Grasses | 58.04 | **59.59** | 56.46 | **59.69** | 54.85 | **58.76** |
| | Amaranth | 74.63 | **87.16** | 79.80 | **88.41** | 69.27 | **81.32** |
| | Goosefoot | 70.07 | **80.79** | 69.75 | **79.99** | 63.72 | **79.95** |
| | Knotweed | 54.82 | **67.93** | 49.72 | **63.46** | 46.57 | **59.91** |
| | Corn spurry | 55.70 | **56.31** | 53.84 | **58.13** | 50.40 | **55.22** |
| | Chickweed | 48.60 | **56.74** | 48.04 | **56.22** | 49.29 | **55.40** |
| | Solanales | 75.48 | **79.34** | 75.93 | **79.94** | 64.59 | **79.77** |
| | Potato weed | 62.13 | **73.42** | 59.92 | **70.21** | 57.87 | **67.79** |
| | Chamomile | 72.27 | **77.70** | 72.94 | **77.18** | 72.86 | **77.23** |
| | Thistle | 71.22 | **75.76** | 63.23 | **75.13** | 62.94 | **69.87** |
| | Mercuries | 6.39 | **9.96** | 5.77 | **9.63** | 3.62 | **8.28** |
| | Geranium | 49.90 | **55.63** | 48.16 | **54.57** | 44.20 | **51.30** |
| | Crucifer | 36.34 | **51.99** | 35.27 | **47.12** | 36.42 | **38.13** |
| | Poppy | **49.65** | 37.58 | 37.04 | **42.88** | **39.55** | 30.94 |
| | Plantago | 45.84 | **55.60** | 40.74 | **56.76** | 32.50 | **46.26** |
| | Labiate | 21.43 | **36.63** | 7.96 | **39.13** | 20.25 | **49.28** |
| mIoU | | 64.45 | **69.99** | 62.35 | **70.01** | 60.09 | **67.74** |
| Class Acc | Background | 99.69 | **99.72** | 99.70 | **99.71** | 99.69 | **99.70** |
| | Maize | **93.74** | 94.73 | 92.96 | **94.77** | 92.77 | **93.69** |
| | Sugar beet | 86.74 | **92.94** | 92.88 | **94.05** | 84.40 | **93.20** |
| | Soy | 90.33 | **91.13** | 89.38 | **91.28** | 87.56 | **91.89** |
| | Sunflower | 96.98 | **97.09** | 96.14 | **97.02** | 95.99 | **96.82** |
| | Potato | 96.88 | **96.99** | 97.09 | **97.19** | **96.88** | 96.58 |
| | Pea | **88.66** | 88.02 | 86.91 | **87.58** | **87.50** | 86.84 |
| | Bean | 87.03 | **88.14** | 83.00 | **91.25** | 82.62 | **87.80** |
| | Pumpkin | 93.56 | **93.79** | 95.83 | **96.21** | 94.93 | **95.72** |
| | Grasses | 77.00 | **78.84** | 75.66 | **78.44** | 73.77 | **77.03** |
| | Amaranth | 89.90 | **92.02** | 88.10 | **93.02** | 82.72 | **91.12** |
| | Goosefoot | 92.07 | **92.88** | 89.94 | **92.60** | 90.95 | **91.14** |
| | Knotweed | 71.65 | **82.21** | 66.83 | **76.75** | 58.46 | **73.08** |
| | Corn spurry | **78.23** | 76.24 | 73.70 | **77.31** | 74.25 | **75.77** |
| | Chickweed | 62.79 | **64.50** | 58.55 | **67.40** | 55.42 | **66.22** |
| | Solanales | 80.54 | **88.67** | 80.33 | **87.90** | 72.01 | **86.46** |
| | Potato weed | 76.17 | **86.94** | 77.02 | **86.11** | 76.97 | **86.06** |
| | Chamomile | **89.85** | 88.46 | **89.37** | 88.90 | 89.31 | **89.93** |
| | Thistle | 82.86 | **86.76** | 74.74 | **86.70** | 72.55 | **83.90** |
| | Mercuries | 7.52 | **11.78** | 6.97 | **11.76** | 4.03 | **10.03** |
| | Geranium | 68.75 | **74.38** | 63.63 | **70.71** | 68.78 | **71.39** |
| | Crucifer | 51.66 | **65.38** | 49.46 | **62.79** | 49.69 | **55.50** |
| | Poppy | **78.27** | 78.00 | 76.08 | **76.96** | **75.43** | 63.37 |
| | Plantago | 56.49 | **72.99** | 60.99 | **73.80** | 41.71 | **60.06** |
| | Labiate | 23.73 | **49.90** | 8.39 | **44.53** | 24.00 | **56.54** |
| mAcc | | 76.84 | **81.30** | 74.94 | **80.99** | 73.30 | **79.19** |

## Supplementary Figures

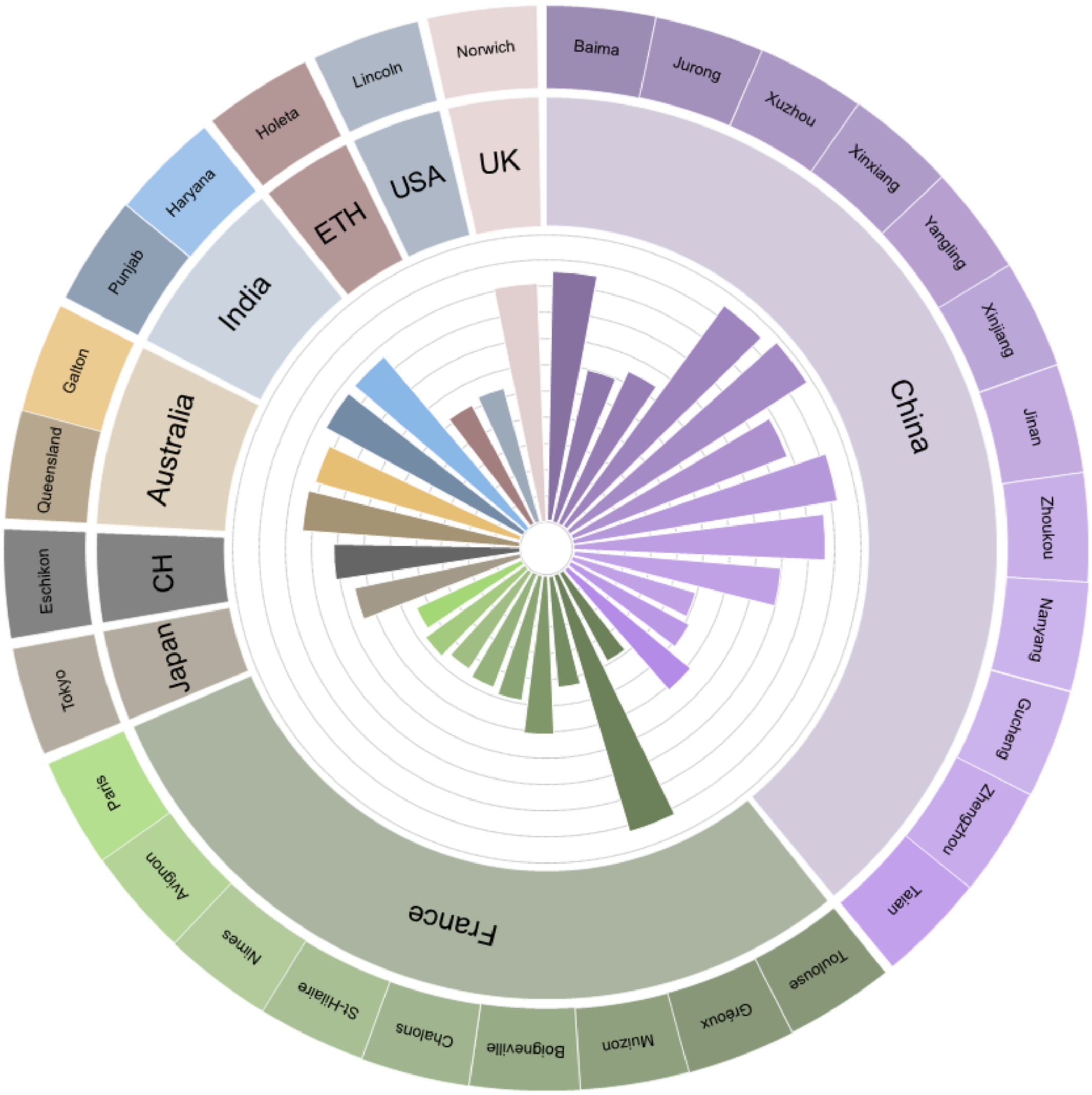

**Supplementary Figure 5.** The dataset comprises image data collected from 29 regions spanning 9 countries, with each region containing multi-year observations. The inner bar chart represents the quantitative distribution of regions per country.

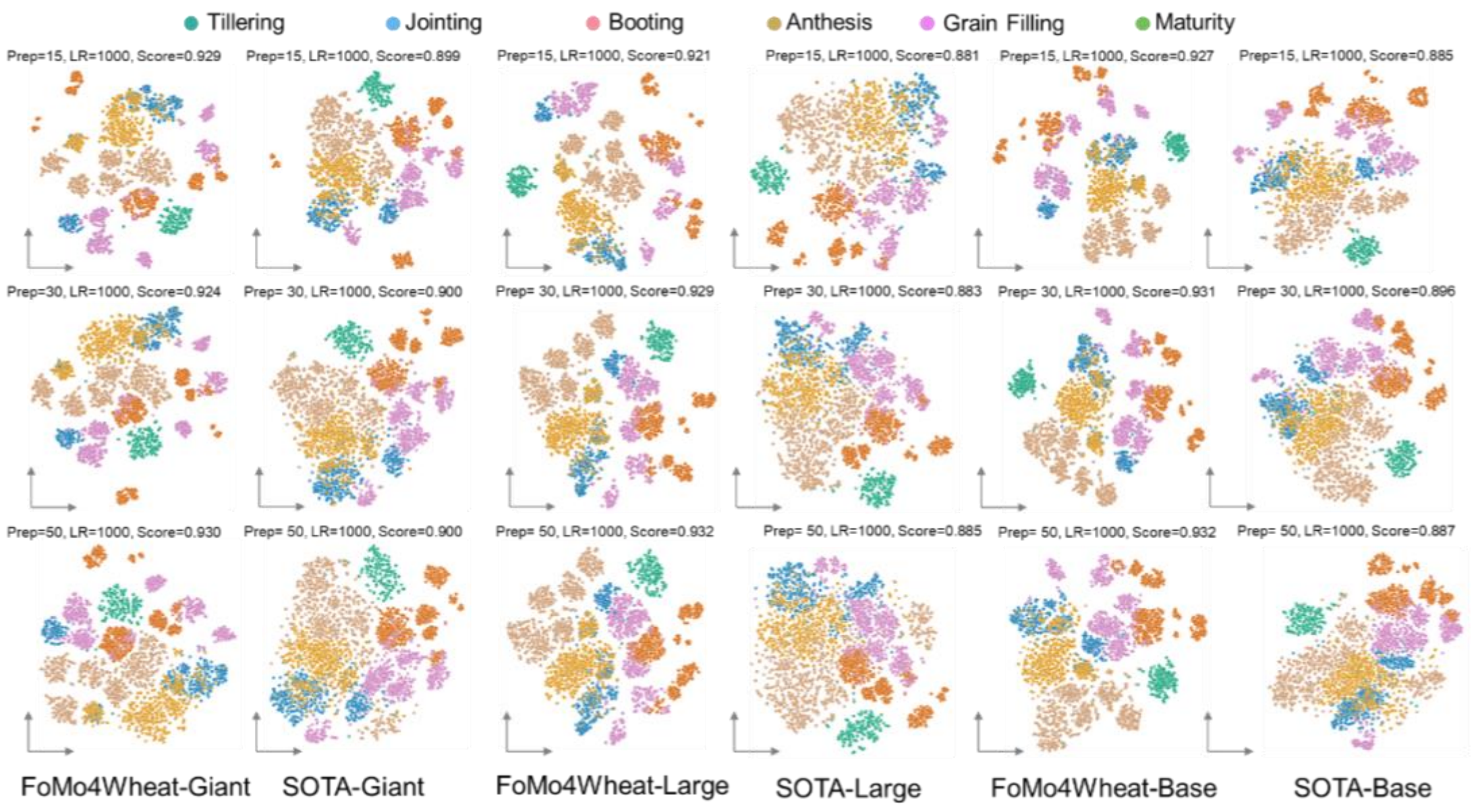

**Supplementary Figure 6. Visual comparison of feature distribution generated by the FoMo4Wheat model and SOTA models on growth stage classification.** Features extracted from different model were visualized using t-SNE with varying perplexity (Prep) settings applied. The x- and y-axes correspond to the first and second dimensions of the reduced space, respectively. To emphasize inter-class separability in the feature space, cross-validated classification accuracy scores were computed using a K-nearest neighbors (KNN) classifier on the 2D t-SNE projections. Notably, the FoMo4Wheat model series consistently outperformed SOTA benchmarks in classification scores.

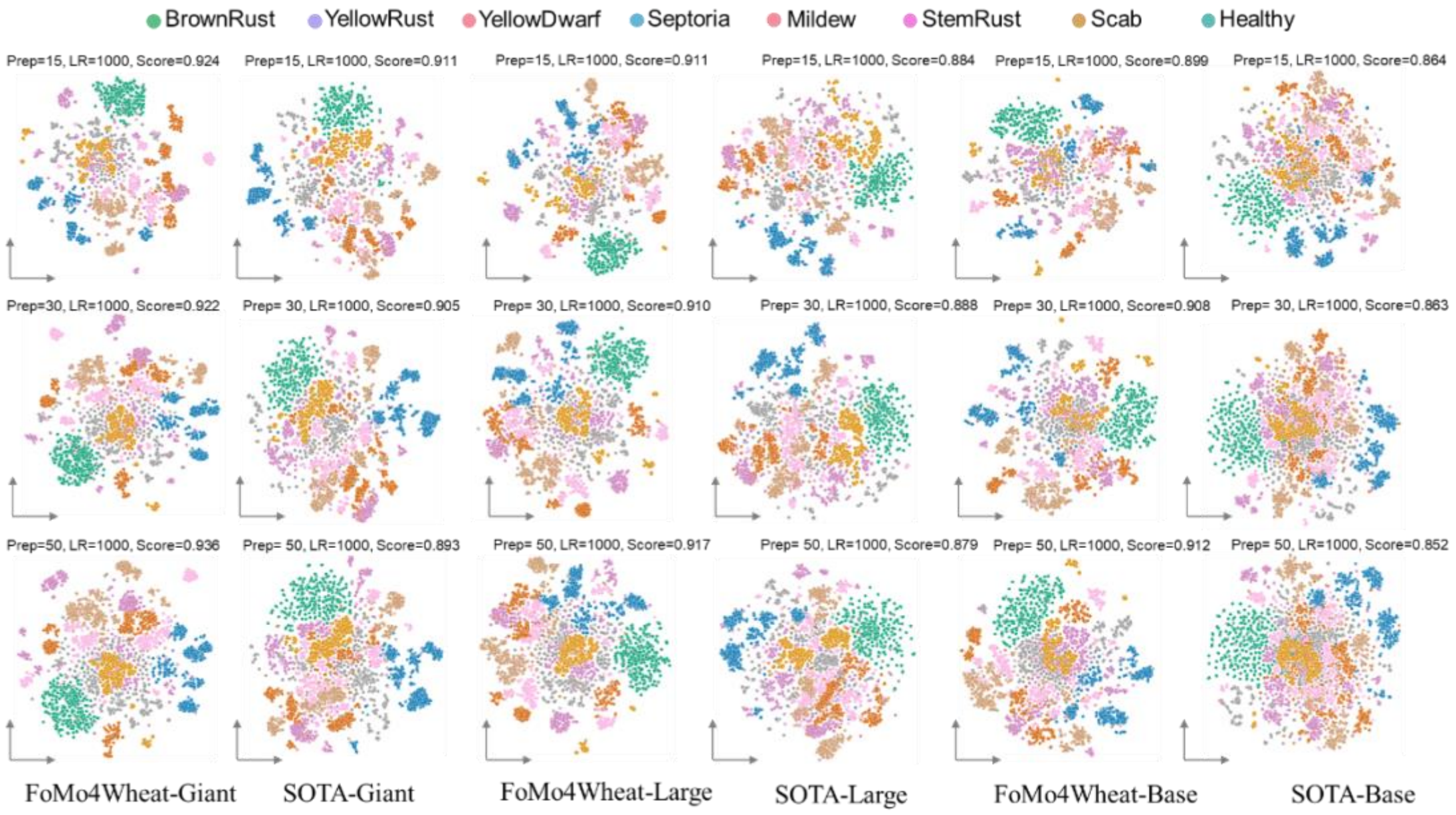

**Supplementary Figure 7. Visual comparison of feature distribution generated by the FoMo4Wheat model and SOTA models on wheat disease classification.** Notably, with a similar visualization and validation protocol as growth stage classification, the FoMo4Wheat model series consistently outperform SOTA models in classification scores.

## Supplementary Recerence

1. Glorot, X. & Bengio, Y. Understanding the difficulty of training deep feedforward neural networks. in *Proceedings of the Thirteenth International Conference on Artificial Intelligence and Statistics* 249–256 (JMLR Workshop and Conference Proceedings, 2010).
2. Shen, L. *et al.* GSP-AI: An AI-Powered Platform for Identifying Key Growth Stages and the Vegetative-to-Reproductive Transition in Wheat Using Trilateral Drone Imagery and Meteorological Data. *Plant Phenomics* **6**, 0255 (2024).
3. Long, M., Hartley, M., Morris, R. J. & Brown, J. K. M. Classification of wheat diseases using deep learning networks with field and glasshouse images. *Plant Pathol.* **72**, 536–547 (2023).
4. Genaev, M. A. *et al.* Image-based wheat fungi diseases identification by deep learning. *Plants* **10**, 1500 (2021).
5. Safarijalal, B., Alborzi, Y. & Najafi, E. Automated Wheat Disease Detection using a ROS-based Autonomous Guided UAV. Preprint at http://arxiv.org/abs/2206.15042 (2022).
6. Mao, R. *et al.* DAE-Mask: a novel deep-learning-based automatic detection model for in-field wheat diseases. *Precis. Agric.* **25**, 785–810 (2024).
7. Goyal, L., Sharma, C. M., Singh, A. & Singh, P. K. Leaf and spike wheat disease detection & classification using an improved deep convolutional architecture. *Inform. Med. Unlocked* **25**, 100642 (2021).
8. He, K., Gkioxari, G., Dollár, P. & Girshick, R. Mask R-CNN. Preprint at https://doi.org/10.48550/arXiv.1703.06870 (2018).
9. Chen, Z. *et al.* Vision Transformer Adapter for Dense Predictions. Preprint at https://doi.org/10.48550/arXiv.2205.08534 (2023).

10. David, E. *et al.* Global Wheat Head Detection 2021: An Improved Dataset for Benchmarking Wheat Head Detection Methods. *Plant Phenomics* **2021**, 2021/9846158 (2021).

11. Liu, C., Lu, H., Cao, Z. & Liu, T. Point-Query Quadtree for Crowd Counting, Localization, and More. in *2023 IEEE/CVF International Conference on Computer Vision (ICCV)* 1676–1685 (IEEE, Paris, France, 2023). doi:10.1109/iccv51070.2023.00161.

12. Li, Y. *et al.* Self-Supervised Plant Phenotyping by Combining Domain Adaptation with 3D Plant Model Simulations: Application to Wheat Leaf Counting at Seedling Stage. *Plant Phenomics* **5**, 0041 (2023).

13. Cheng, B., Misra, I., Schwing, A. G., Kirillov, A. & Girdhar, R. Masked-attention Mask Transformer for Universal Image Segmentation. in *2022 IEEE/CVF Conference on Computer Vision and Pattern Recognition (CVPR)* 1280–1289 (IEEE, New Orleans, LA, USA, 2022). doi:10.1109/cvpr52688.2022.00135.

14. Wang, Z. *et al.* The Global Wheat Full Semantic Organ Segmentation (GWFSS) Dataset. 2025.03.18.642594 Preprint at https://doi.org/10.1101/2025.03.18.642594 (2025).

15. Zhou, J. *et al.* Global Rice Multiclass Segmentation Dataset (RiceSEG): Comprehensive and Diverse High-Resolution RGB-Annotated Images for the Development and Benchmarking of Rice Segmentation Algorithms. *Plant Phenomics* 100099 (2025) doi:10.1016/j.plaphe.2025.100099.

16. Madec, S. *et al.* VegAnn, Vegetation Annotation of multi-crop RGB images acquired under diverse conditions for segmentation. *Sci. Data* **10**, 302 (2023).

17. Steininger, D., Trondl, A., Croonen, G., Simon, J. & Widhalm, V. The CropAndWeed Dataset: a Multi-Modal Learning Approach for Efficient Crop and Weed Manipulation. in *2023 IEEE/CVF Winter Conference on Applications of Computer Vision (WACV)* 3718–3727 (IEEE, Waikoloa, HI, USA, 2023).

doi:10.1109/WACV56688.2023.00372.

18. Wheat Leaf dataset. https://www.kaggle.com/datasets/olyadgetch/wheat-leaf-dataset.

19. CGIAR Wheat Growth Stage Challenge. https://www.kaggle.com/datasets/gauravduttakiit/cgiar-wheat-growth-stage-challenge.